\documentclass[letterpaper]{article} 
\usepackage{aaai25}  
\usepackage{times}  
\usepackage{helvet}  
\usepackage{courier}  
\usepackage[hyphens]{url}  
\usepackage{graphicx} 
\usepackage{natbib}  
\usepackage{caption} 
\usepackage{algorithm}
\usepackage{algorithmic}

\usepackage{url}            
\usepackage{booktabs}       
\usepackage{amsfonts}       
\usepackage{nicefrac}       
\usepackage{microtype}      
\usepackage{xcolor}         
\usepackage{graphicx}
\usepackage{subfigure}
\usepackage{caption}
\usepackage{array}
\usepackage{float}
\usepackage{amsmath}
\usepackage{subcaption}
\usepackage{booktabs}
\usepackage{multirow}

\usepackage{amsmath,amsfonts,bm}

\def\eqref#1{equation~\ref{#1}}

\def\1{\bm{1}}

\DeclareMathAlphabet{\mathsfit}{\encodingdefault}{\sfdefault}{m}{sl}
\SetMathAlphabet{\mathsfit}{bold}{\encodingdefault}{\sfdefault}{bx}{n}

\usepackage{scalerel}
\usepackage{xr}
\newcommand\Tau{\scalerel*{\tau}{T}}
\usepackage{url}
\usepackage{booktabs} 

\usepackage{newfloat}
\usepackage{listings}
\DeclareCaptionStyle{ruled}{labelfont=normalfont,labelsep=colon,strut=off} 
\floatstyle{ruled}
\newfloat{listing}{tb}{lst}{}
\floatname{listing}{Listing}

\nocopyright 

\title{Efficient Symmetry-Aware Materials Generation \\ via Hierarchical Generative Flow Networks}
\author{
    Tri Minh Nguyen, 
    Sherif Abdulkader Tawfik, 
    Truyen Tran,
    Sunil Gupta, 
    Santu Rana, 
    Svetha Venkatesh
}
\affiliations{
    Applied Artificial Intelligence Institute\\
    Deakin University\\
    Victoria, Australia
}

\usepackage{bibentry}

\begin{document}

\maketitle

\begin{abstract}
Discovering new solid-state materials requires rapidly exploring the vast space of crystal structures and locating stable regions. Generating stable materials with desired properties and compositions is extremely difficult as we search for very small isolated pockets in the exponentially many possibilities, considering elements from the periodic table and their 3D arrangements in crystal lattices. Materials discovery necessitates both optimized solution structures and diversity in the generated material structures. Existing methods struggle to explore large material spaces and generate diverse samples with desired properties and requirements.
We propose the Symmetry-aware Hierarchical Architecture for Flow-based Traversal (SHAFT), a novel generative model employing a hierarchical exploration strategy to efficiently exploit the symmetry of the materials space to generate crystal structures given desired properties. In particular, our model decomposes the exponentially large materials space into a hierarchy of subspaces consisting of symmetric space groups, lattice parameters, and atoms.
We demonstrate that SHAFT significantly outperforms state-of-the-art iterative generative methods, such as Generative Flow Networks (GFlowNets) and Crystal Diffusion Variational AutoEncoders (CDVAE), in crystal structure generation tasks, achieving higher validity, diversity, and stability of generated structures optimized for target properties and requirements.
\end{abstract}

%

\section{Introduction}
\label{intro}

Discovering new solid-state materials plays a central role in advancing multiple critical industries, including energy generation and storage, and semiconductor electronics \citep{Berger2020SemiconductorMaterials,Noh2019InverseRepresentation}.  Each unique crystal structure exhibits properties useful for specific applications. For example, superconductive perovskite structure is used in circuit board elements for computers.
Generating material structures that meet given property requirements poses a set of unique challenges.
The key challenge is generating crystal structures that exhibit a repeating arrangement of atoms in three-dimensional space. A crystal structure is determined by how atoms are arranged within the unit cell specified by their lengths and angles. However, the inter-atom interactions are not confined within the unit cell but also with adjacent unit cells. These characteristics make the search space of crystal structures significantly larger and more complex compared to well-studied molecular search space. The number of known crystal structures, both experimental and hypothetical, is around 3 million curated from AFlow \citep{Mehl2017The1,Hicks2019The2,Hicks2021The3} and Material Project \citep{Jain2013Commentary:Innovation}, which is tiny compared to billions of molecules from the Zinc dataset \citep{Irwin2005ZINC-AScreening}. The limited data undermines modern data-driven methods to learn the crystal structure representation \citep{Chithrananda2020ChemBERTa:Prediction,Liu2019RoBERTa:Approach}.

\begin{figure}[ht]
\begin{center}
\includegraphics[scale=0.45]{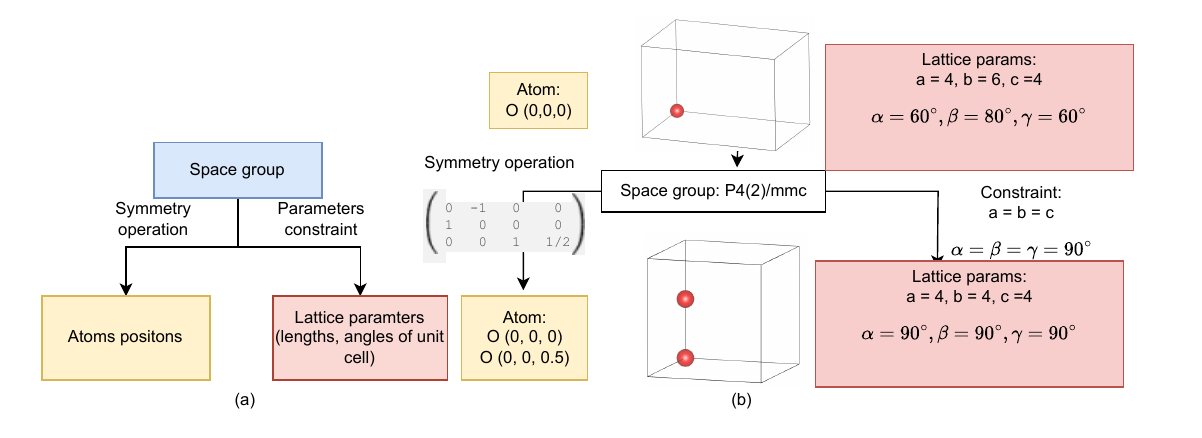}
\end{center}
\caption{(a) Hierarchical crystal structure state. The space group level provides the set of symmetry operations for atoms' positions and lattice parameters constraints. (b) An example of applying the hierarchical state space. The current state has one Oxygen atom at position (0, 0, 0), lattice parameters $a=4,b=6,c=4,\alpha=60^\circ, \beta=80^\circ, \gamma=60^\circ$, and $P1$ spacegroup. The action of choosing space group P4(2)/mmc provides a symmetry operation to generate another Oxygen atom at position (0, 0, 0.5). The lattice parameter constraints reduce the unit cell's length search space from $\mathbb{R}^3$ to $\mathbb{R}$ and make the unit cell's angles constant at $90^\circ$}
\label{fig:hstate}
\end{figure}

We address the complexity associated with the large search space by proposing a new generative model termed Hierarchical Symmetry-aware Hierarchical Architecture for Flow-based Traversal (SHAFT).  SHAFT explores the vast search space in an efficient way. The key insight to solving the large state space problem is breaking space exploration into more meaningful hierarchical sub-tasks. Here the higher-level tasks explore more actions that are closely related to the reward function while lower-level tasks handle the configuration adjustment corresponding to the action taken at higher-level tasks. Since the exploration starts at the highly general concept level, it can learn a more meaningful policy that corresponds to the target reward function. With more meaningful actions taken at a higher level, the policy networks can focus on searching the actions in a significantly smaller sub-space that corresponds to actions of high-level tasks instead of exploring the whole space.

Three key concepts can help generative models search the material space effectively. First, the crystal structure class imposes symmetry operations and geometrical characteristics on atom positions and lattice parameters, as seen in Fig.~\ref{fig:hstate}, effectively reducing the material search space.
Second, searching for stable structures on the non-smooth energy landscape defined by quantum mechanics requires generative models to explore and generate diverse sample sets. This approach helps avoid getting stuck in a single mode or local minima. Third, we apply bond constraints on atom pairs in the generated crystals, ensuring atoms are not closer to each other than specific bond thresholds. These thresholds are obtained using all materials in the MaterialsProject database. Given the diversity of materials in this database, it's reasonable to assume that minimum distances derived from it are threshold distances for the ground state structure of the respective atoms in any new crystal structure. By incorporating these concepts, generative models can more effectively navigate the complex material space and produce stable, diverse, and physically realistic crystal structures.


We model the entire material space in a hierarchical structure. The highest level is the crystal structure class, which is related to relevant material properties. Exploring this class can efficiently lead to more optimal properties of the generated structure. Specifically, we use the space group of the crystal structure to effectively reduce the 3D atom space exploration to a more meaningful high-level exploration and generate a high-symmetry crystal structure. The next level searches for the unit cell lattice parameters and atom configurations given the space group. The space group imposes constraints on these parameters and atom positions, thus reducing the search space. Choosing positions one by one creates a long horizon trajectory, making it difficult to learn a policy that generates high-symmetry structures proportional to the reward function.
By leveraging the space group's symmetry operations, we can immediately replicate atoms over the unit cell. This approach reduces the trajectory length, making it easier for the policy network to learn how to generate a high-symmetry crystal structure.

In this work, we apply our proposed SHAFT to the materials structure generation. Our main contributions are:

$>$We propose a generative model that can search effectively in a large search space by modeling the state space in a hierarchical structure.

$>$We incorporate the physical knowledge curated from the large material databases into our generative model to generate more stable structures.

$>$We validate the hierarchical structure state space and physics priors of our proposed generative model in the crystal structure generation task to show the efficiency in material space exploration as well as the stability of the generated structures.

\section{Related Works}
\label{relatedwork}

There are three main approaches to crystal structure generation. $>$The first is \textbf{Element Substitution}: Given a crystal template \citep{Belkly2002NewDesign}, the elements are substituted with other elements having similar properties \citep{Hautier2011DataCompounds,Wang2021PredictingSimilarity,Wei2022TCSP:Discovery}. This approach is computationally expensive and relies on domain knowledge in element substitution.  
$>$The second strategy is \textbf{Distribution Sampling}: Deep generative models learn from the distribution of known stable crystal structures to sample new structures, such as those by Variational AutoEncoders (VAEs) \citep{Xie2022CrystalGeneration,Court20203-DLearning,Noh2019InverseRepresentation}, Generative Adversarial Networks (GANs) \citep{Zhao2021High-ThroughputNetworks,Kim2020GenerativePrediction},  symmetry-aware Diffusion Models \citep{Luo2023TowardsMaterials}, and periodic-E(3)-equivariant denoising model \citep{Jiao2023CrystalDiffusion}. This data-driven approach has difficulty in generating high symmetry structures and in out-of-distribution generalization. Also, the generation process is solely based on learning from the known data, leaving minimal room for domain knowledge and human intervention during the generation process.  
And finally, $>$\textbf{Iterative Generation}: Crystal structures can be decomposed into compositional objects, and constructed step by step using reinforcement learning (RL) \citep{Zamaraeva2023ReinforcementPrediction}. Crystal-GFN \citep{AI4Science2023Crystal-GFN:Constraints} uses GFlownet to generate crystal structures' parameters such as space group, lattice lengths, and angles, but it does not generate the complete structure with atoms' coordinates. Our work is the first to apply the RL-based technique to explore the entire material and atom space. The key advantage of this RL-based approach is that it allows the incorporation of domain knowledge into the action and state space and shaping the reward function. The RL can generalise to \textbf{Hierarchical RL} which solves the long horizon trajectory problem by designing sub-goals \citep{Nachum2018Data-EfficientLearning,WenDeepMind2020OnLearning}. By exploiting high-level crystal structure classes such as space groups, SHAFT creates a sub-goal for low-level policies, helping them to learn the atom coordinates space more effectively.     

\textbf{Generative Flow Networks} (GFlowNet) are generative models designed to sample candidates proportional to their target rewards. The framework has been successfully applied to many fields such as molecular discovery \citep{Bengio2021FlowGeneration}, protein sequence discovery \citep{Jain2022BiologicalGFlowNets}, causality \citep{Deleu2022BayesianNetworks}, and continuous control \citep{Li2023CFlowNets:Networks}. Jiangyan et al. \citep{JiangyanMa2023BakingGFlowNets} integrate symmetries into GFlowNets with equivalent actions. LS-GFN \citep{MinsuKim2024LocalGFlowNets} explores the local neighborhood to avoid the over-exploration in GFlownet. Our SHAFT is a generalization of GFlowNets for more efficient sampling by modelling the hierarchy of the search space. When applied to crystal structure generation, SHAFT also exploits the group symmetry unique to this domain.

\section{Preliminaries}
\label{preliminaries}


\subsection{Crystallographic Space Group}
\label{Subsec:spacegroup}


The crystal structure is the repeating arrangement of the atoms within a unit cell in 3D.  Formally, the unit cell of $N$ atoms is a triplet $(L, A, X)$ of lattice parameters $L$, atom list $A$, and atom coordinates $X$. There are 6 lattice parameters $L=(a, b, c, \alpha, \beta, \gamma) \in \mathbb{R}^6$ describing 3 lengths and 3 angles of the unit cell, respectively. The atoms list $A = (a_1,...,a_N)$ describes the elements. The atoms' coordinates $X \in \mathbb{R}^{N\times 3}$ describe the positions of the atoms within the unit cell, which can be Cartesian or fractional.


The space group of crystal structure consists of a list of symmetry transformations to the atoms within the unit cell. In crystallography, there are 230 space groups \citep{Glazer2012SpaceScientists}.
Each space group has geometrical characteristics defined in lattice angles and lengths that can be used as constraints to limit the parameters search space. The list of geometrical characteristics is provided in the Appendix \ref{sec:lattice_spacegroup_character}. Given the space group $G_s$, the elements of the space group $g \in G_s$ are a set of symmetry operations. A crystallographic orbit of an atom $o=(x_o,a_o)$ with coordinate $x_o$ and element $a_o$ is defined as 
$O_{G_s}(x_o) = \{g \cdot x_o \mid g \in G_s\}$,
where $g \cdot x_o$ denotes the application of the symmetry operation $g$ on the atom $o$ within the unit cell. From a reference atom $o$, we can obtain a set $O_{G_s}$ of equivalent points.

\subsection{Generative Flow Network}
GFlowNet models the sampling process of the compositional object $s$  as the directed acrylic graph (DAG) $G = (S,A)$ where $S$ is the set of states and $A$ is the state transition which is the subset of $\mathcal{S}\times\mathcal{S}$. The sampling process starts with the initial state vertex $s_0 \in S$ with no incoming edge. and stops at the sink state vertex $s_n \in S$, $n$ is the sampling trajectory length, with no outgoing edge. GFlowNet learns a policy function $\pi$ that can sample the object $x$ with the probability proportional to the non-negative reward function. GFlowNets constructs the object step by step, from the initial state $s_0$ to the sink state $s_n$, forming a trajectory $\tau={s_0,..,s_n}, \tau \in \Tau$ where $\Tau$ is the trajectory set. For any state $s$ in the trajectory, we can define the flow of the state as $F(s) = \sum_{\tau \ni s}F(\tau)$ and the flow of the edge $s \rightarrow s'$ as $F(s) = \sum_{\tau \ni s \rightarrow s'}F(\tau)$ \citep{Bengio2021FlowGeneration}. The forward policy that maps the probability of transition from the current state $s$ to the next state $s'$ is given as $P_F(s'|s)=\frac{F(s\rightarrow s')}{F(s)}$. The backward policy mapping probability transition to the previous state $s$ given the current state $s'$ is $P_B(s|s')=\frac{F(s\rightarrow s')}{F(s')}$. The training objective of the GFlowNets is flow matching consistency where the incoming flow is equal to the outgoing flow, $\sum_{s''\rightarrow s}F(s''\rightarrow s) = F(s) = \sum_{s\rightarrow s'}F(s\rightarrow s')$, for all states. \cite{Malkin2022TrajectoryGFlowNets} propose trajectory balance to deal with the long trajectory credit assignment problem.  Given the trajectory $\tau = (s_0 \rightarrow s_1 \rightarrow ... \rightarrow s_n)$, the forward probability of a trajectory is defined as $\prod_{t=1}^n P(s'|s)$. The trajectory balance constraint is defined as:
\begin{equation}
\label{eq:tb_constraint}
    Z \prod_{t=1}^n P_F(s_t|s_{t-1}) = F(x) \prod_{t=1}^n P_B(s_{t-1}|s_t) 
\end{equation} 
where $P(s_n = x) = \frac{F(x)}{Z}$, $Z = F(s_0)$ is the initial state flow.
Then the trajectory balance objective is defined as:
\begin{equation}
\label{eq:TB_obj}
    \mathcal{L}_{TB}(\tau) = \left(\log\frac{Z_{\theta}\prod_{t=1}^n P_F(s_t|s_{t-1};\theta)}{ R(x) \prod_{t=1}^n P_B(s_{t-1}|s_t;\theta) }\right)^2
\end{equation}


\section{Methods}
\label{sec:hstatespace}

To best leverage the group symmetry in the crystal structure space, here we propose SHAFT (Symmetry-aware Hierarchical Architecture for Flow-based Traversal), a generic flow-based generative model that explores the space in a hierarchical manner. In SHAFT, the lower states represent discrete concepts constrained by the higher states that represent more abstract concepts. When applied to crystal structure generation, SHAFT makes use of the group symmetry in the space. See Fig. \ref{fig:hstate}  for the overall structural state design.

\subsection{Hierarchical Policy}
\label{sec:hpolicy}




\begin{figure}[h]
\begin{center}
\includegraphics[scale=0.6]{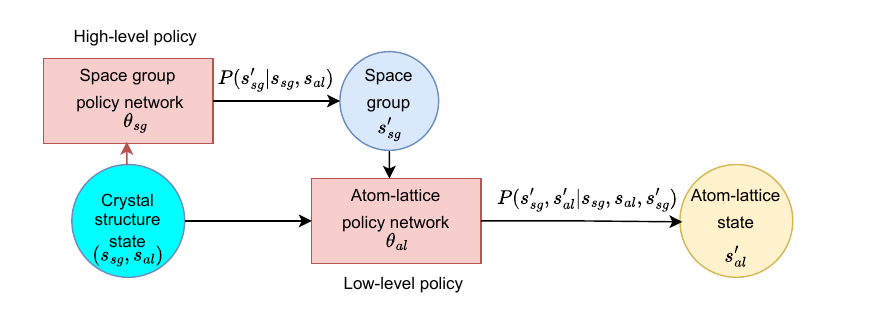}
\end{center}
\caption{Hierarchical policy for crystal structure state. First, the crystal structure graph state $s$ is decomposed into space group state $s_{sg}$ and atom-lattice state $s_{al}$. Then the transition probability $P(s'_{sg}|s_{sg},s_{al})$ in Eq. \ref{eq:transist_prob} is given by the space group policy network $\theta_{sg}$. The transition probability $P(s'_{al}|s_{sg},s_{al},s'_{sg})$ in Eq. \ref{eq:transist_prob} is given by the atom-lattice policy network $\theta_{al}$.}
\label{fig:hspolicy}
\end{figure}

SHAFT generalises GFlowNets to accommodate a policy hierarchy of two levels: The high-level decision-making policy operating on the space groups, and the low-level execution policy operating on the atom-lattices (see Fig.~\ref{fig:hspolicy}). The space group policy chooses the space group and applies corresponding constraints on the atom-lattice policy actions.  The corresponding hierarchical state space is decomposed as $s=(s_{sg},s_{al})$, where $s_{sg}$ is the space group state and $s_{al}$ is the atom-lattice state. The latter consists of lattice parameters $s_{lp}$, atoms' coordinate $s_{ac}$, and atoms' type $s_{at}$ states.  

Then we define the probability of transitions as:
\begin{align}
    P(s'|s) &= P(s'_{sg},s'_{al}|s_{sg},s_{al}) \\
    P(s'_{sg},s'_{al}|s_{sg},s_{al}) &=  P(s'_{al}|s_{sg},s_{al},s'_{sg})P(s'_{sg}|s_{sg},s_{al}) \label{eq:transist_prob}
\end{align} 
Then given the trajectory $\tau = (s_0 \rightarrow s_1 \rightarrow ... \rightarrow s_n)$, we then have the trajectory balance constraint (Eq. \ref{eq:tb_constraint}) with the decomposed state as:
\begin{align}
\label{eq:hier_tb_constraint}
    Z \prod_{t=1}^n P_F(s^t_{al}|s^{t-1}_{sg},s^{t-1}_{al},s^t_{sg})P_F(s^t_{sg}|s^{t-1}_{sg},s^{t-1}_{al})\\ 
    = F(x) \prod_{t=1}^n P_B(s^{t-1}_{al}|s^{t}_{sg},s^{t}_{al},s^{t-1}_{sg})P_B(s^{t-1}_{sg}|s^{t}_{sg},s^{t}_{al})
\end{align} 
The space group forward and backward transition probabilities $P_F(s_{sg}^t|s^{t-1})$ and $P_B(s_{sg}^{t-1}|s^{t})$ are parameterized by the multinoulli distribution defined by the logits output of the space group policy networks. The lattice parameters state $s_{lp}$ transition probability is parameterized by the  Gaussian distribution defined by the mean $\mu$ and variance $\sigma^2$. The atom fraction coordinates state $s_{ac}$ transition probability is parameterized by the  Multivariate Gaussian distribution given by the mean $\mu$ and covariance matrix $\Sigma$. The atom type state $s_{at}$ transition probability is parameterized by the multinoulli distribution.


The lattice sampling process starts with the state $s_0$ which has lattice parameters as $a=b=c=4, \alpha=\beta=\gamma=90^o$, space group $1$, empty crystal graph at state $s_0$ $\mathcal{G}_{s_0}$. The sampling process is described in Algorithm~\ref{alg:sampling}.

\begin{algorithm}[tb]
   \caption{Hierarchical Trajectory Sampling}
   \label{alg:sampling}
\begin{algorithmic}
   \STATE {\bfseries Input:} $\theta_{sg}$, $\theta_{al}$, T, $min_l$, $max_l$
   \STATE {\bfseries Return:} Trajectory $\tau$, complete crystal structure $x$
    \STATE {\bfseries Initialize: }
    
    \STATE  $s_{al} \gets \{a=b=c=4, \alpha=\beta=\gamma=90^o\}$\;

    \STATE $s_{sg} \gets 1$
    
    \STATE $\mathcal{G} \gets \oslash$
    
    Reference atom list $A_{ref} \gets \oslash$
   \FOR{step $t=1$ {\bfseries to} $T$}
   \STATE Set $p_{nsg} \gets \theta_{sg}(s_{sg}, \mathcal{G})$.\\
   \STATE Space group $s'_{sg} \sim M_{nsg}(1,p_{nsg})$.\\
   \STATE Set $\alpha_{lattice}$, $\beta_{lattice}, \alpha_{atom}$, $\beta_{atom}$, $p_{ne}$ $\gets$ $\theta_{sg}(s'_{sg}, s_{sg}$, $\mathcal{G})$\\
   \STATE Lattice parameter  $a$, $b$, $c$, $\alpha$, $\beta$, $\gamma$ $\sim$ $Beta$$(\alpha_{lattice}$, $\beta_{lattice})$
   \STATE An atom's fraction coordinate $(x, y, z) \sim Beta(\alpha_{atom}, \beta_{atom})$\\
   \STATE Get the atom's element valid $mask$ based on the composition constraint.
   \STATE Set $p_{ne}[mask] = -\infty$.
   \STATE Atom's element $s'_{at} \sim M_{nsg}(1,p_{ne})$\\
   \STATE Set $A_{ref} = A_{ref} \cup (s'_{at},s'_{ac},)$ \\
    \STATE Add state $s'_{sg}, s'_{lp}, s'_{at}, s'_{ac}$ to trajectory $\tau$\\
   
   \ENDFOR
\end{algorithmic}
\end{algorithm}

\subsection{Physics-informed Reward Function}
\label{sec:reward}


At the terminal state $x = s_{M_t}$ of the trajectory, a reward is returned by a non-negative function $R(x)$, providing feedback on the generated crystal structure, especially its validity and stability. To achieve these goals, we design the reward function to include the following key elements:

The \textbf{formation energy term} dictates that a stable crystal structure will have a negative formation energy, thus defined as $R_e(x) = e^{-E(x)/T}$
where $E(x)$ is the predicted formation energy per atom given by the prediction model (Appendix Sec. \ref{sec:formemodel}), $T$ is the normalization term.

The \textbf{bond length preferences term} $R_b$ encourages the bond lengths of the input crystal structure to agree with the empirical lengths retrieved from the database. First, for each atom $a_i$ in the crystal structure $x$, we search all its neighbors $nei(a_i)$ with the radius cutoff $4.0$ \r{A}. The neighbors can also include atoms from adjacent unit cells. We retrieve the minimum and average bond distance of all bond types existing in $x$ from the Materials Project database (e.g. minimum bond distance of Li-O is 1.63 and the average bond distance of Li-O is 3.02). Then the difference between the bond length $d(a_i, a_j), a_j \in nei(a_i)$ with minimum $d_{min}(a_i,a_j)$ and average bond distance $d_{avg}(a_i,a_j)$ of $(a_i, a_j)$ bond type.
\begin{equation}
\begin{split}
    R_{bond}(x) = & \frac{|d(a_i,a_j)-d_{avg}(a_i,a_j)|}{n_{bond}} \\ & + e^{2(d_{min}(a_i,a_j)-d(a_i,a_j))} \\ \forall a_i \in x, &\forall a_j \in nei(a_i)
\end{split}
\end{equation}
where $n_{bond}$ is the total number bond in the crystal structure $x$. We then regularize the bond term $R_{bond}(x)$ to $(0,1)$ as $R_{bond}(x) = e^{-R_{bond}(x)}$. 

The \textbf{density term} $R_P$ encourages the structure density. Because the bond distance preference term applies a strict penalty for structures violating the minimum distance constraint, the generative model tends to generate a structure with long distances between atom pairs with only a few neighbor atoms. This leads the model to generate gas with low density rather than solid-state density. On the other hand, structures with very high density (i.e. larger than 10) are unlikely to be realistic as the units are crowded with atoms causing a very high formation energy. We measure the generated structure density $P(x)$ (Appendix Sec. \ref{sec:density}) and define the density term as the Gaussian function of structure density:
$R_P(x) = ae^{\frac{-(P(x)-b)^2}{2c^2}}$.


The \textbf{composition validity term} is defined as $R_{comp}(x)=1$ for valid composition and $R_{comp}(x)=0$ otherwise. We follow composition validity given by the charge neutrality check \citep{Davies2019SMACT:Theory}. Finally, the \textbf{physic-informed reward function} is composed as:
\begin{equation}
    \label{eq:reward-func}
    R(x) = w_eR_e(x) + w_pR_P(x) + w_bR_{bond}(x) + w_cR_{comp}(x)
\end{equation}
where $w_e, w_p, w_b, w_c$ are weighted scalar. We further discuss these terms in Appendix  Sec.\ref{sec:motivationrewardfunction}  

\subsection{State Representation}
\label{sec:state_encoding}


The crystal structure state must capture all relevant information about the crystal, including its atomic arrangement and the lattice parameters. We represent the crystal unit cell $(L, A, X)$ described in Sec.~\ref{Subsec:spacegroup} as a directed graph $\mathcal{G} = (\mathcal{V}, \mathcal{E})$ of node feature matrix $\mathcal{V}$ and edge matrix $\mathcal{E}$. The node features include the atom's atomic number and its fractional coordinates within the unit cell. The edges are determined by k-nearest neighbor with the maximum number of neighbors being 12 and the radius cut-off is $8.0$ \r{A}.
The graph representation of the crystal structure is learned using a MatErials Graph Network (MEGNet) model \cite{Chen2019GraphCrystals} (details in Appendix Sec. \ref{sec:megnet}) to satisfy the periodic invariance and E(3) invariance of crystal structure.  


Lattice parameters are encoded using a multi-layer perceptron as:
\begin{equation}
\begin{split}
    h_{\mathcal{L}} = &\text{MLP}([l_1, l_2, l_3, \sin(\alpha), \cos(\alpha), \sin(\beta), \cos(\beta), \\
    & \sin(\gamma),\cos(\gamma)]) 
\end{split}
\end{equation}
where $l_1$, $l_2$, $l_3$ are the lattice lengths, and $\alpha, \beta, \gamma$ are the angle of the lattice angle. The space group is encoded as $h_{sg}$ using an embedding layer. Finally, the crystal structure state is simply $h_M = (h_{sg}, s_{al})$ where $h_{al} = [h_{\mathcal{G}};h_{\mathcal{L}}]$. The encoded crystal structure state $h_M$ is later used as the representation of the hierarchical state space $s_M$. 




\section{Experiments}
\label{sec:exp}

Our SHAFT can be applied to any crystal discovery tasks given the set of elements. The main bottleneck is the prohibitive cost of validation of the generated structures. We will first focus on a constrained generation task with careful DFT validation (Sec.~\ref{sec:battery}). Then we demonstrate the scalability of SHAFT on nearly unconstrained generation tasks (Sec.~\ref{sec:largespace})

\textbf{Baselines}
We compare our SHAFT with SOTA techniques including CDVAE \citep{Xie2022CrystalGeneration} and GFlowNets, which is a flat version of our method. CDVAE is a diffusion model that uses SE(3) equivariant GNNs adapted with periodicity. We train CDVAE on the MP-Battery dataset for battery material discovery task (based on requirements in Sec.\ref{sec:battery}, more details in Appendix Sec. \ref{sec:mpbatterydataset}) and the MP-20 dataset \cite{Jain2013Commentary:Innovation} for general crystal generation (Sec. \ref{sec:largespace}).
We follow the recent work on continuous GFlownet \citep{Lahlou2023ANetworks} to work on the continuous space of the atoms' coordinates and lattice parameters. The model has a single-level policy network that outputs space group, lattice parameters, atoms' coordinates, and atoms' type. Note that this is the first time GFlowNet has been applied successfully to complete crystal generation.

\subsection{Constrained Generation: Battery Materials Discovery}
\label{sec:battery}

For concreteness, we focus on battery material discovery, the most fundamental task in the battery industry worth hundreds of billions of dollars, and of great importance in securing a green energy future. The challenge is to generate stable structures with specific formulas and elements.

Motivated by the search for light-weight, transition-metal free cation materials, we explore the space of possible materials that can be made from the light elements Be, B, C, N, O, Si, P, S, and Cl, and one of the three alkali metals Li, Na and K. This space of materials constitutes materials that can be utilised in lithium-ion, sodium-ion, and potassium-ion battery materials, respectively.
We discuss further the motivation of the experiments in Appendix~\ref{sec:batterydiscovery}. 

\begin{figure}[ht]
\begin{center}
\includegraphics[scale=0.3]{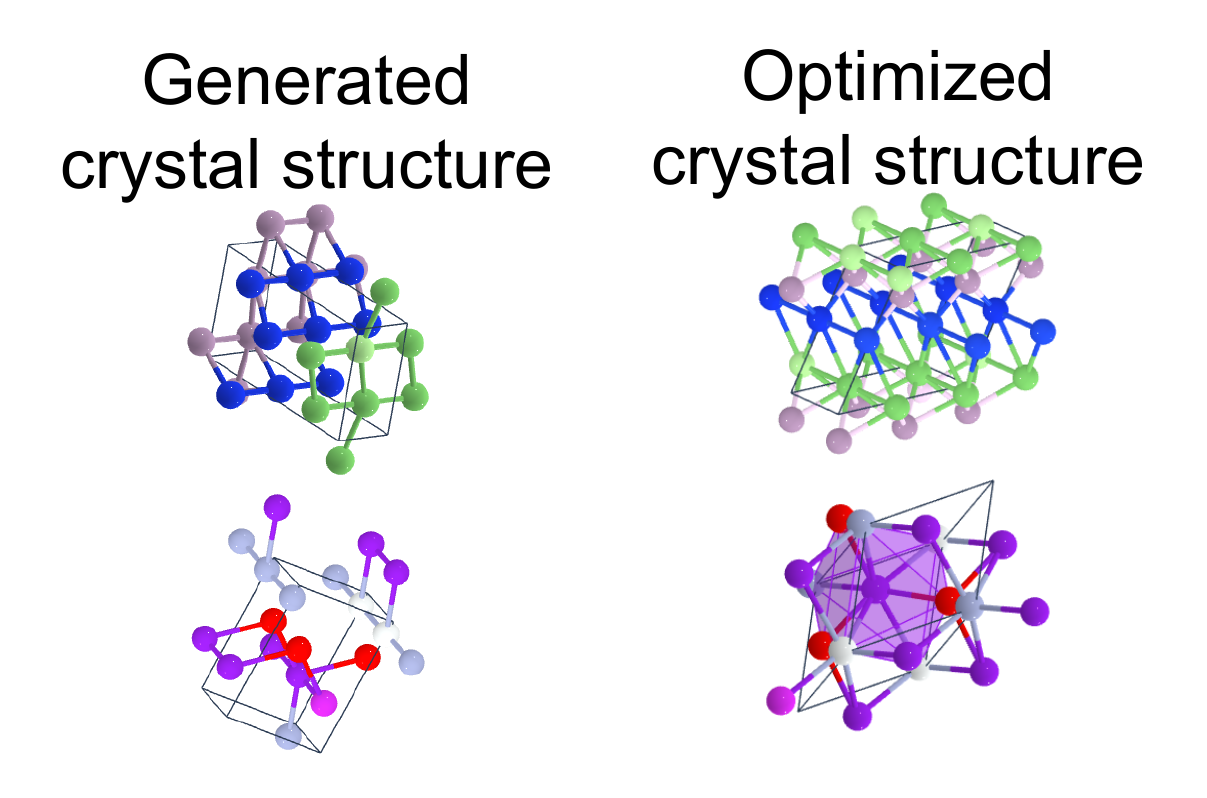}
\end{center}
\caption{Examples of generated crystal structures and the corresponding structure optimized by M3GNet framework \citep{Chen2022ATable}. }
\label{fig:gen_examples}
\end{figure}

\begin{table}
\caption{Validity of the generated structures. We evaluate the top-K crystal structures ranked by the reward function after $8\times10^5$ states visited.}
\label{Tab1:valid_3}
\begin{center}
\begin{tabular}{m{1.8cm}m{1.8cm}m{2.1cm}}
\hline
 Method & Structure $\uparrow$ & Composition $\uparrow$\\ \hline
CDVAE       &0.66$\pm 0.03$  &0.73$\pm 0.09$ \\
GFlowNet    &0.98$\pm 0.03$   &1.00$\pm 0.00$  \\
SHAFT   &0.99$\pm 0.01$   &1.00$\pm 0.00$ \\
\hline
\end{tabular}
\end{center}
\end{table}





 \begin{table}
\caption{Diversity of the generated structures. We evaluate the top-K crystal structures ranked by the reward function after $8\times10^5$ states visited.}
\label{Tab1:diverse_3}
\begin{center}
\begin{tabular}{m{1.7cm} m{1.2cm} m{1.2cm}m{1.2cm}}
\hline
Method & Crystal family $\uparrow$& Compo- sition $\uparrow$ & Structure $\uparrow$ \\ \hline
CDVAE       & 0.15 & 1859.07 & 0.43\\
&$\pm0.09$&$\pm199.28$&$\pm0.004$\\
GFlowNet    & 1.97 & 165.13 & 0.84 \\
&$\pm0.15$&$\pm32.55$&$\pm0.03$\\
SHAFT   & 2.19& 147.80 & 0.89 \\
&$\pm0.06$ &$\pm15.81$&$\pm0.05$\\
\hline
\end{tabular}
\end{center}
\end{table}
\paragraph{Material Validity}


 We evaluate the proposed method and baseline methods on the validity of the generated crystal structure, measured based on two criteria.  We follow the previous work \citep{Xie2022CrystalGeneration} for validity measurements a) \textit{structure validity}: a structure is valid as long as the minimum distance between any two atoms is more than $0.5$ \r{A} b) \textit{composition validity}: a composition is valid if the overall charge computed by SMACT \cite{Davies2019SMACT:Theory} is neutral. As seen in Tab.~\ref{Tab1:valid_3}, both GFlowNet and SHAFT structure and composition validities are close to one, highlighting the effectiveness of learning reward-based exploration. The structure and composition validities are used in the reward function described in Sec.~\ref{sec:reward}.  In contrast, CDVAE faces the common problem of sampling from data-induced distribution which is low structure validity. 
\begin{table}
\caption{Average of formation energy and $E_{hull}$ of the generated structures. We evaluate the top-K crystal structures ranked by the reward function after $8\times10^5$ states visited.}
\label{Tab1:forme_3}
\begin{center}
\begin{tabular}{p{1.5cm}p{2cm}p{1.5cm}l}
\hline
Method & Energy $\downarrow$ & \% energy $< 0$ $\uparrow$ & $E_{hull}$ $\downarrow$\\ \hline
CDVAE &-0.318$\pm0.080$    &74.6$\pm8.6$ & 1.04 $\pm0.24$\\
GFlowNet &-1.014$\pm0.003$&98.6$\pm1.6$ & 0.32 $\pm0.02$\\
SHAFT  &-1.241$\pm0.097$&99.5$\pm0.8$  & 0.29 $\pm0.03$\\
\hline
\end{tabular}
\end{center}
\end{table}
\paragraph{Material Diversity}

Following the previous works \citep{Xie2022CrystalGeneration,Zhao2023PhysicsConstraints}, we evaluate both the structure and composition diversity of the generated crystal structures. The \textit{structure diversity} is defined as the average pairwise Euclidean distance between the structure fingerprint of any two generated materials \citep{Pan2021BenchmarkingStructures}. The \textit{composition diversity} is defined as the average pairwise distance between the composition fingerprints of any two generated materials \citep{Pan2021BenchmarkingStructures}. More details are provided in \ref{sec:diversity}. We further use the crystal family defined in Supplement Tab. \ref{stab:lattice_spacegroup} which is a group of space groups sharing some special geometric characteristics. The \textit{crystal family diversity} is defined as the  Shannon–Wiener index \citep{Shannon1948ACommunication} of the number of generated structures in each crystal family. We evaluate the diversity of structures discovered by the models and report the results in Tab.~\ref{Tab1:diverse_3}.

\begin{figure}
\begin{center}
{\includegraphics[width=\columnwidth]{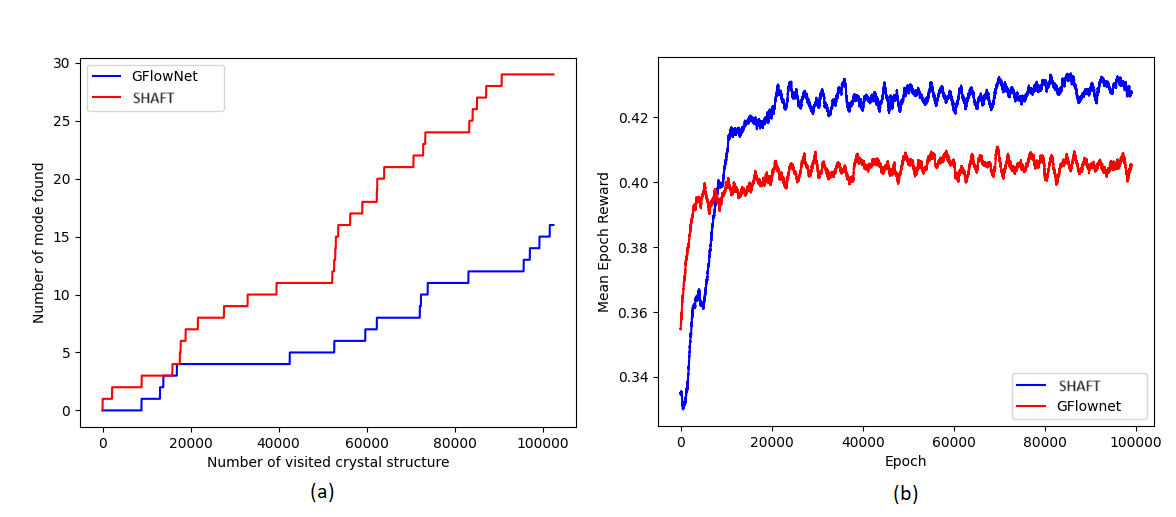}}
\end{center}
\caption{(a) Comparison of SHAFT and GFlowNet in exploring crystal modes using 3 steps. A mode is defined as a valid crystal structure with negative formation energy. A step is an action of choosing one atom in the spacegroup-lattice-atom hierarchical state space. (b) The average reward of crystal structure sampled by GFlownet and SHAFT in each epoch.} \label{fig:mode_found}
\end{figure}
\paragraph{Formation Energy} 
We also report the average formation energy and the percentage of structure with formation energy per atom smaller than $0$ eV/atom, and energy above the hull. 
Compared with the flat GFlowNets, our SHAFT find crystals with lower formation energy and more diversity in terms of crystal family and structure.

\paragraph{Modes Exploration}
To evaluate the speed of exploring the material space and finding the valid material structures, we count the number of modes found by the generated crystal structures plotted against the number of states visited in Fig.~\ref{fig:mode_found}a. We define a unique \textit{mode} as a valid crystal structure satisfying conditions (Appendix Sec. \ref{sec:mode}). The results show that using our hierarchical model improves the speed of mode discovery compared to the flat variant. The average reward per epoch in Fig.\ref{fig:mode_found}b shows that SHAFT not only speeds up the mode discovery but also discovers higher reward structures. The formation energy, bond length preferences, density, and composition validity term improving over epochs (see Fig.\ref{fig:term_ep}) shows that the model can learn to embed the physical constraints into the policy network. Fig.\ref{fig:term_ep_dist} showing the shift of the distribution of reward function terms from random crystal structure to trained SHAFT further proves this point.  


\paragraph{Stability of Generated Materials}

\begin{table}
\caption{Match rates of the generated crystal paired with structures optimized by M3GNet \citep{Chen2022ATable}}
\label{tab:matchM3GNet}
\begin{center}
\begin{tabular}{p{2cm}p{2cm}p{3cm}}
\toprule
Methods &Match rate $\uparrow$ & RMS displacement $\downarrow$ \\
\hline
CDVAE & 0.46 $\pm0.03$ & 0.287 $\pm0.0133$\\
GFlowNet & 0.80 $\pm0.01$ & 0.174 $\pm0.019$ \\
SHAFT  & 0.82 $\pm0.02$ & 0.161 $\pm0.020$\\
\bottomrule
\end{tabular}
\end{center}
\end{table}

It is a common practice to relax the generated crystal structures to seek the lower potential energy surface using DFT calculation iteratively. As DFT calculation is expensive, it is desirable to generate a structure that is close to energy minima. In this experiment, we compare the generated crystal structure with its optimized structure. We use the M3GNet framework \citep{Chen2022ATable} to iteratively optimize the energy predicted by the potential surface energy model. Examples of generated structures and their corresponding optimized structures are shown in Fig. \ref{fig:gen_examples}.

\textbf{Match rate} We follow the previous work in \citep{Zhao2023PhysicsConstraints} to evaluate the match rate of crystal structure relaxation. A structure $m$ and its optimized structure $m'$ are matched if their atoms' translation and angle are within tolerance thresholds, indicating that the generated structure is close to the optimal, and thus is more stable. We use the matching algorithm provided by pymatgen library \citep{Ong2013PythonAnalysis} in StructureMacher with $10^o$ angle tolerance, $1.0$ fractional length tolerance, and $1.0$ site tolerance. The match rate is the fraction of the number of matched structures on the total number of generated structures.  

\begin{figure}
\begin{centering}
{\includegraphics[width=0.7\columnwidth]{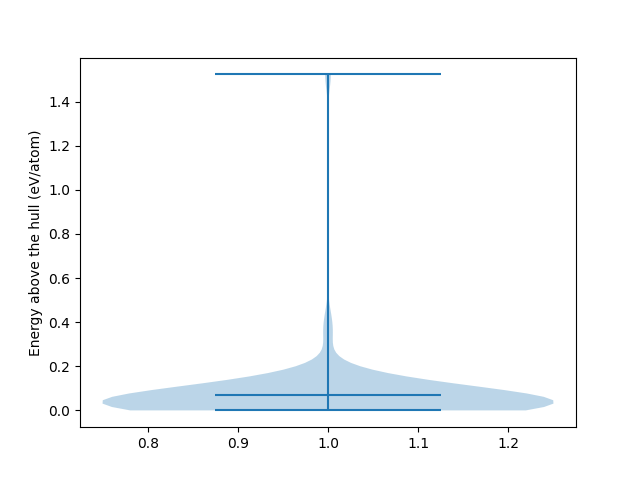}}
\par\end{centering}
\caption{The distribution of energy above the hull of DFT optimized structures. Most of the generated are stable ($E_{hull}=0 eV/atom$) or metastable $E_{hull}<0.05 eV/atom$. We can achieve low $E_{hull}$ without optimizing directly.} \label{fig:dft_dist}
\end{figure}
The results reported in Tab.~\ref{tab:matchM3GNet} show that our proposed method can produce more structures (at the rate of 0.82) that are nearly optimal in terms of total energy compared to GFlowNets and CDVAE.

\paragraph{DFT Validation}

We sample and choose the top-K generated materials for DFT verification. Out of 100 crystal structures, 95\% (95) are successfully optimized. The complete list of DFT-optimized structures is included in the Supplement material. Fig.\ref{fig:dft_dist} shows that most DFT-optimized structures are stable or metastable ($E_{hull}<0.05 eV/atom$). The generated and optimized structures with their total energy per atom, $E_{hull}$, and further results and calculation details are available in the Appendix.

\paragraph{Ablation Study}
To demonstrate the ability to guide the generative model to generate more stable structures, we perform the ablation study on the reward function's terms of Eq.~\ref{eq:reward-func}.
The generated crystal structures are relaxed using M3GNet optimization framework \citep{Chen2022ATable}. In addition, we also compare GCN policy network (details in Appendix Sec.\ref{sec:gcnmp}) and MEGNet policy network (Appendix Sec. \ref{sec:state_encoding},\ref{sec:megnet}).
\begin{table}[h]
\caption{The ablation study on the impact of reward function terms on the crystal structure stability. Match rates of the generated crystal paired with structures optimized by M3GNet \citep{Chen2022ATable}}
\label{tab:matchM3GNet_ablation}
\begin{center}
\begin{tabular}{p{1.5cm}p{0.7cm}p{0.7cm}p{0.743cm}p{0.7cm}p{0.7cm}p{0.7cm}}
\toprule
Methods & Comp. val. & Struct. val. & Avg. Form. E. & \% E. $< 0$ &Match rate $\uparrow$ & RMS dis. $\downarrow$\\
\hline
SHAFT  (MEGNet)         &1.00&0.99&-1.17&100&0.82&0.16\\
W/o density    &1.00&1.00&-1.20&100&0.76&0.16\\
W/o bond score  &0.99&0.32&-1.43&100&0.53&0.19 \\
W/o formation energy &1.00&1.00&-0.99&96.0&0.77&0.17 \\
W/o validity &1.00&1.00&-1.15&100&0.72&0.17\\
SHAFT  (GCN) &1.00&0.98&-1.15&100&0.81&0.16\\
\bottomrule
\end{tabular}
\end{center}
\end{table}
The results reported in Tab.~\ref{tab:matchM3GNet_ablation} show that all the density, bond score, and formation energy terms are necessary for the model to generate more stable structures. Surprisingly, without the formation energy training signal, the model still can produce structures with negative formation energy with atoms being placed at their preferred distances (distilled from large datasets such as Material Project) combined with appropriate density. 




The result also shows that it is desirable to have an expressive GNN as a policy network as it can capture the permutation, translation, rotation, and periodic invariance of crystal structure and learn more accurate structure representation.

\paragraph{Complexity Analysis}
\label{sec:runtime}

Our experiments are conducted on a single GPU Tesla V100-SXM2-32GB with 80 cores CPU Intel(R) Xeon(R) CPU E5-2698 v4 @ 2.20GHz. We report the average training time per iteration, sampling time per sample, and the number of parameters of policy networks (including backward and forward policy) of our proposed method using the reported configuration and a sampling-based method (CDVAE) in Tab. \ref{tab:runtime}.

We follow the crystal structure graph construction from Matformer \cite{Yan2022PeriodicPrediction}. According to complexity analysis \cite{Yan2022PeriodicPrediction}, assuming each node has at least 12 neighbors, the constructed graph $G=(V,E)$ has $|V|=n$ and $|E|=6n$. In case of high multiplicity of high symmetry or low-density crystal structure, the number of edges to evaluate can be very high.    

\begin{table}[h]
\caption{The average training time per iteration, sampling per sample, and the number of parameters of policy networks of our proposed method using the reported configuration and a sampling-based method (CDVAE)}
\label{tab:runtime}
\begin{center}
\begin{tabular}{lccc}
\toprule
Methods & Time/epoch  & Sampling & Model Params. \\
\hline
CDVAE & 3.28s & 1.97s & 4.9M\\ 
SHAFT  & 0.89s & 0.061s & 0.9M\\
\bottomrule
\end{tabular}
\end{center}
\end{table}

\subsection{Exploring a Very Large Element Space}
\label{sec:largespace}
To further demonstrate the ability to generalize to the whole chemical space, we train our proposed method on 89 elements from H to Pu. Because we have 89 instead of 12 elements, the composition diversity is higher than the experiment in Tab. \ref{Tab1:diverse_3}. As shown in Tab.\ref{tab:general_chem_space} and \ref{tab:general_chem_space_diverse}, our proposed method maintains a high percentage of stable structure and low formation energy.


\begin{table}[ht]
\caption{The validity, energy, and stability experiment results of CDVAE, GFlownet and SHAFT on 89 elements}
\label{tab:general_chem_space}
\begin{center}
\begin{tabular}{p{1.5cm}p{0.7cm}p{0.7cm}p{0.75cm}p{0.7cm}p{0.7cm}p{0.7cm}}
\toprule
Methods & Comp. val. & Struct. val. & Avg. Form. E. & \% E. $< 0$ &Match rate $\uparrow$ & RMS dis. $\downarrow$\\
\hline
CDVAE &0.86&1.00&-0.84&58& 0.71&0.14\\
GFlownet &1.00&1.00&-0.89&72&0.66&0.18\\
SHAFT &1.00&0.97&-1.14&87&0.71&0.19\\
\bottomrule
\end{tabular}
\end{center}
\end{table}

\begin{table}[h]
\caption{The diversity experiment results of CDVAE, GFlownet, and SHAFT on 89 elements}
\label{tab:general_chem_space_diverse}
\begin{center}
\begin{tabular}{lp{2cm}p{2cm}p{2cm}}
\toprule
Methods & Crystal family diversity $\uparrow$& Composition diversity $\uparrow$& Structure diversity $\uparrow$\\
\hline
CDVAE & 0.95&1857.31&0.74\\
GFlownet &2.37&1399.59&0.99\\
SHAFT &2.32&1471.59&1.01\\
\bottomrule
\end{tabular}
\end{center}
\end{table}

\section{Conclusion}

We have proposed SHAFT, a Hierarchical Generative Flow Network for crystal structure generation, aiming at rapid exploration of the exponential crystal space and simultaneously satisfying physics constraints. SHAFT is built on a hierarchical state space, allowing for multi-level policy networks to operate on action abstraction. It effectively exploits the high-symmetry in the crystal structure space, defining space transformation groups. The framework is flexible for domain experts to embed domain knowledge to guide the generation process through space structure design and reward engineering. SHAFT demonstrates its superiority in efficiency in exploration, diversity, and stability in the generated crystal structures. 
While our focus is on materials discovery, our hierarchical state space and policy can easily extend to other tasks such as continuous control.

\bibliography{references}

\begin{thebibliography}{50}
\providecommand{\natexlab}[1]{#1}

\bibitem[{AI4Science et~al.(2023)AI4Science, Hernandez-Garcia, Duval, Volokhova, Bengio, Sharma, Carrier, Benabed, Koziarski, and Schmidt}]{AI4Science2023Crystal-GFN:Constraints}
AI4Science, M.; Hernandez-Garcia, A.; Duval, A.; Volokhova, A.; Bengio, Y.; Sharma, D.; Carrier, P.~L.; Benabed, Y.; Koziarski, M.; and Schmidt, V. 2023.
\newblock {Crystal-GFN: sampling crystals with desirable properties and constraints}.

\bibitem[{Belkly et~al.(2002)Belkly, Helderman, Karen, and Ulkch}]{Belkly2002NewDesign}
Belkly, A.; Helderman, M.; Karen, V.~L.; and Ulkch, P. 2002.
\newblock {New developments in the Inorganic Crystal Structure Database (ICSD): accessibility in support of materials research and design}.
\newblock \emph{Acta crystallographica. Section B, Structural science}, 58(Pt 3 Pt 1): 364--369.

\bibitem[{Bengio et~al.(2021)Bengio, Jain, Korablyov, Precup, and Bengio}]{Bengio2021FlowGeneration}
Bengio, E.; Jain, M.; Korablyov, M.; Precup, D.; and Bengio, Y. 2021.
\newblock {Flow Network based Generative Models for Non-Iterative Diverse Candidate Generation}.
\newblock \emph{Advances in Neural Information Processing Systems}, 33: 27381--27394.

\bibitem[{Berger(2020)}]{Berger2020SemiconductorMaterials}
Berger, L.~I. 2020.
\newblock \emph{{Semiconductor Materials}}.
\newblock CRC Press.
\newblock ISBN 9780138739966.

\bibitem[{Chen and Ong(2022)}]{Chen2022ATable}
Chen, C.; and Ong, S.~P. 2022.
\newblock {A universal graph deep learning interatomic potential for the periodic table}.
\newblock \emph{Nature Computational Science 2022 2:11}, 2(11): 718--728.

\bibitem[{Chen et~al.(2019)Chen, Ye, Zuo, Zheng, and Ong}]{Chen2019GraphCrystals}
Chen, C.; Ye, W.; Zuo, Y.; Zheng, C.; and Ong, S.~P. 2019.
\newblock {Graph Networks as a Universal Machine Learning Framework for Molecules and Crystals}.
\newblock \emph{Chemistry of Materials}, 31(9): 3564--3572.

\bibitem[{Chithrananda, Grand, and Ramsundar(2020)}]{Chithrananda2020ChemBERTa:Prediction}
Chithrananda, S.; Grand, G.; and Ramsundar, B. 2020.
\newblock {ChemBERTa: Large-scale self-supervised pretraining for molecular property prediction}.
\newblock In \emph{Machine Learning for Molecules Workshop, NeurIPS}. Online.

\bibitem[{Court et~al.(2020)Court, Yildirim, Jain, and Cole}]{Court20203-DLearning}
Court, C.~J.; Yildirim, B.; Jain, A.; and Cole, J.~M. 2020.
\newblock {3-D inorganic crystal structure generation and property prediction via representation learning}.
\newblock \emph{Journal of Chemical Information and Modeling}, 60(10): 4518--4535.

\bibitem[{Davies et~al.(2019)Davies, Butler, Jackson, Skelton, Morita, and Walsh}]{Davies2019SMACT:Theory}
Davies, D.; Butler, K.; Jackson, A.; Skelton, J.; Morita, K.; and Walsh, A. 2019.
\newblock {SMACT: Semiconducting Materials by Analogy and Chemical Theory}.
\newblock \emph{Journal of Open Source Software}, 4(38): 1361.

\bibitem[{Deleu et~al.(2022)Deleu, G{\'{o}}is, Emezue, Rankawat, Lacoste-Julien, Bauer, and Bengio}]{Deleu2022BayesianNetworks}
Deleu, T.; G{\'{o}}is, A.; Emezue, C.; Rankawat, M.; Lacoste-Julien, S.; Bauer, S.; and Bengio, Y. 2022.
\newblock {Bayesian Structure Learning with Generative Flow Networks}.
\newblock In \emph{Proceedings of the Conference on Uncertainty in Artificial Intelligence}.

\bibitem[{Glazer, Burns, and Glazer(2012)}]{Glazer2012SpaceScientists}
Glazer, M.; Burns, G.; and Glazer, A.~N. 2012.
\newblock \emph{{Space groups for solid state scientists}}.

\bibitem[{Hautier et~al.(2011)Hautier, Fischer, Ehrlacher, Jain, and Ceder}]{Hautier2011DataCompounds}
Hautier, G.; Fischer, C.; Ehrlacher, V.; Jain, A.; and Ceder, G. 2011.
\newblock {Data mined ionic substitutions for the discovery of new compounds}.
\newblock \emph{Inorganic Chemistry}, 50(2): 656--663.

\bibitem[{Hicks et~al.(2021)Hicks, Mehl, Esters, Oses, Levy, Hart, Toher, and Curtarolo}]{Hicks2021The3}
Hicks, D.; Mehl, M.~J.; Esters, M.; Oses, C.; Levy, O.; Hart, G.~L.; Toher, C.; and Curtarolo, S. 2021.
\newblock {The AFLOW Library of Crystallographic Prototypes: Part 3}.
\newblock \emph{Computational Materials Science}, 199: 110450.

\bibitem[{Hicks et~al.(2019)Hicks, Mehl, Gossett, Toher, Levy, Hanson, Hart, and Curtarolo}]{Hicks2019The2}
Hicks, D.; Mehl, M.~J.; Gossett, E.; Toher, C.; Levy, O.; Hanson, R.~M.; Hart, G.; and Curtarolo, S. 2019.
\newblock {The AFLOW Library of Crystallographic Prototypes: Part 2}.
\newblock \emph{Computational Materials Science}, 161: S1--S1011.

\bibitem[{Irwin and Shoichet(2005)}]{Irwin2005ZINC-AScreening}
Irwin, J.~J.; and Shoichet, B.~K. 2005.
\newblock {ZINC-A free database of commercially available compounds for virtual screening}.
\newblock \emph{Journal of Chemical Information and Modeling}, 45(1): 177--182.

\bibitem[{Jain et~al.(2013)Jain, Ong, Hautier, Chen, Richards, Dacek, Cholia, Gunter, Skinner, Ceder, and Persson}]{Jain2013Commentary:Innovation}
Jain, A.; Ong, S.~P.; Hautier, G.; Chen, W.; Richards, W.~D.; Dacek, S.; Cholia, S.; Gunter, D.; Skinner, D.; Ceder, G.; and Persson, K.~A. 2013.
\newblock {Commentary: The Materials Project: A materials genome approach to accelerating materials innovation}.
\newblock \emph{APL Materials}, 1(1).

\bibitem[{Jain et~al.(2022)Jain, Bengio, Garcia, Rector-Brooks, Dossou, Ekbote, Fu, Zhang, Kilgour, Zhang, Simine, Das, and Bengio}]{Jain2022BiologicalGFlowNets}
Jain, M.; Bengio, E.; Garcia, A.-H.; Rector-Brooks, J.; Dossou, B. F.~P.; Ekbote, C.; Fu, J.; Zhang, T.; Kilgour, M.; Zhang, D.; Simine, L.; Das, P.; and Bengio, Y. 2022.
\newblock {Biological Sequence Design with GFlowNets}.
\newblock \emph{Proceeding of International Conference on Machine Learning}.

\bibitem[{{Jiangyan Ma} et~al.(2023){Jiangyan Ma}, {Emmanuel Bengio}, {Yoshua Bengio}, and {Dinghuai Zhang}}]{JiangyanMa2023BakingGFlowNets}
{Jiangyan Ma}; {Emmanuel Bengio}; {Yoshua Bengio}; and {Dinghuai Zhang}. 2023.
\newblock {Baking Symmetry into GFlowNets}.
\newblock In \emph{NeurIPS AI for Science Workshop}.

\bibitem[{Jiao et~al.(2023)Jiao, Huang, Lin, Han, Chen, Lu, and Liu}]{Jiao2023CrystalDiffusion}
Jiao, R.; Huang, W.; Lin, P.; Han, J.; Chen, P.; Lu, Y.; and Liu, Y. 2023.
\newblock {Crystal Structure Prediction by Joint Equivariant Diffusion}.
\newblock In \emph{Proceedings of the Advances in Neural Information Processing Systems}.

\bibitem[{Kahle, Marcolongo, and Marzari(2020)}]{Kahle2020High-throughputConductors}
Kahle, L.; Marcolongo, A.; and Marzari, N. 2020.
\newblock {High-throughput computational screening for solid-state Li-ion conductors}.
\newblock \emph{Energy {\&} Environmental Science}, 13(3): 928--948.

\bibitem[{Kim et~al.(2014)Kim, Park, Park, Lim, Hong, and Kang}]{Kim2014NovelBatteries}
Kim, H.; Park, Y.-U.; Park, K.-Y.; Lim, H.-D.; Hong, J.; and Kang, K. 2014.
\newblock {Novel transition-metal-free cathode for high energy and power sodium rechargeable batteries}.
\newblock \emph{Nano Energy}, 4: 97--104.

\bibitem[{Kim et~al.(2020)Kim, Noh, Gu, Aspuru-Guzik, and Jung}]{Kim2020GenerativePrediction}
Kim, S.; Noh, J.; Gu, G.~H.; Aspuru-Guzik, A.; and Jung, Y. 2020.
\newblock {Generative Adversarial Networks for Crystal Structure Prediction}.
\newblock \emph{ACS Central Science}, 6(8): 1412--1420.

\bibitem[{Kipf and Welling(2017)}]{Kipf2017Semi-supervisedNetworks}
Kipf, T.~N.; and Welling, M. 2017.
\newblock {Semi-supervised classification with graph convolutional networks}.
\newblock In \emph{Proceedings of the International Conference on Learning Representations}. Toulon, France.

\bibitem[{Lahlou et~al.(2023)Lahlou, Deleu, Lemos, Zhang, Volokhova, Hern{\'{a}}ndez-Garc{\'{i}}a, N{\'{e}}hale~Ezzine, Bengio, and Malkin}]{Lahlou2023ANetworks}
Lahlou, S.; Deleu, T.; Lemos, P.; Zhang, D.; Volokhova, A.; Hern{\'{a}}ndez-Garc{\'{i}}a, A.; N{\'{e}}hale~Ezzine, L.; Bengio, Y.; and Malkin, N. 2023.
\newblock {A Theory of Continuous Generative Flow Networks}.
\newblock In \emph{Proceedings of the International Conference on Machine Learning}.

\bibitem[{Li et~al.(2023)Li, Luo, Wang, and HAO}]{Li2023CFlowNets:Networks}
Li, Y.; Luo, S.; Wang, H.; and HAO, J. 2023.
\newblock {CFlowNets: Continuous control with Generative Flow Networks}.

\bibitem[{Liu et~al.(2019)Liu, Ott, Goyal, Du, Joshi, Chen, Levy, Lewis, Zettlemoyer, Stoyanov, and Allen}]{Liu2019RoBERTa:Approach}
Liu, Y.; Ott, M.; Goyal, N.; Du, J.; Joshi, M.; Chen, D.; Levy, O.; Lewis, M.; Zettlemoyer, L.; Stoyanov, V.; and Allen, P.~G. 2019.
\newblock {RoBERTa: A robustly optimized BERT pretraining approach}.
\newblock \emph{arXiv preprint arXiv:1907.11692}.

\bibitem[{Luo, Liu, and Ji(2023)}]{Luo2023TowardsMaterials}
Luo, Y.; Liu, C.; and Ji, S. 2023.
\newblock {Towards Symmetry-Aware Generation of Periodic Materials}.
\newblock In \emph{Proceedings of Conference on Neural Information Processing Systems}.

\bibitem[{Malkin et~al.(2022)Malkin, Jain, Bengio, Sun, and Bengio}]{Malkin2022TrajectoryGFlowNets}
Malkin, N.; Jain, M.; Bengio, E.; Sun, C.; and Bengio, Y. 2022.
\newblock {Trajectory balance: Improved credit assignment in GFlowNets}.
\newblock In \emph{Proceedings of Advances in Neural Information Processing Systems}.
\newblock ISBN 9781713871088.

\bibitem[{Mehl et~al.(2017)Mehl, Hicks, Toher, Levy, Hanson, Hart, and Curtarolo}]{Mehl2017The1}
Mehl, M.~J.; Hicks, D.; Toher, C.; Levy, O.; Hanson, R.~M.; Hart, G.; and Curtarolo, S. 2017.
\newblock {The AFLOW Library of Crystallographic Prototypes: Part 1}.
\newblock \emph{Computational Materials Science}, 136: S1--S828.

\bibitem[{{Minsu Kim} et~al.(2024){Minsu Kim}, {Taeyoung Yun}, {Emmanuel Bengio}, {Dinghuai Zhang}, {Yoshua Bengio}, {Sungsoo Ahn}, and {Jinkyoo Park}}]{MinsuKim2024LocalGFlowNets}
{Minsu Kim}; {Taeyoung Yun}; {Emmanuel Bengio}; {Dinghuai Zhang}; {Yoshua Bengio}; {Sungsoo Ahn}; and {Jinkyoo Park}. 2024.
\newblock {Local Search GFlowNets}.
\newblock In \emph{Proceeding of International Conference on Learning Representations}.

\bibitem[{Muy et~al.(2019)Muy, Voss, Schlem, Koerver, Sedlmaier, Maglia, Lamp, Zeier, and Shao-Horn}]{Muy2019High-ThroughputDescriptors}
Muy, S.; Voss, J.; Schlem, R.; Koerver, R.; Sedlmaier, S.~J.; Maglia, F.; Lamp, P.; Zeier, W.~G.; and Shao-Horn, Y. 2019.
\newblock {High-Throughput Screening of Solid-State Li-Ion Conductors Using Lattice-Dynamics Descriptors}.
\newblock \emph{iScience}, 16: 270--282.

\bibitem[{Nachum et~al.(2018)Nachum, Brain, Gu, Lee, and Levine}]{Nachum2018Data-EfficientLearning}
Nachum, O.; Brain, G.; Gu, S.; Lee, H.; and Levine, S. 2018.
\newblock {Data-Efficient Hierarchical Reinforcement Learning}.
\newblock In \emph{Conference on Neural Information Processing Systems}.

\bibitem[{Noh et~al.(2019)Noh, Kim, Stein, Sanchez-Lengeling, Gregoire, Aspuru-Guzik, and Jung}]{Noh2019InverseRepresentation}
Noh, J.; Kim, J.; Stein, H.~S.; Sanchez-Lengeling, B.; Gregoire, J.~M.; Aspuru-Guzik, A.; and Jung, Y. 2019.
\newblock {Inverse Design of Solid-State Materials via a Continuous Representation}.
\newblock \emph{Matter}, 1(5): 1370--1384.

\bibitem[{Ong et~al.(2013)Ong, Richards, Jain, Hautier, Kocher, Cholia, Gunter, Chevrier, Persson, and Ceder}]{Ong2013PythonAnalysis}
Ong, S.~P.; Richards, W.~D.; Jain, A.; Hautier, G.; Kocher, M.; Cholia, S.; Gunter, D.; Chevrier, V.~L.; Persson, K.~A.; and Ceder, G. 2013.
\newblock {Python Materials Genomics (pymatgen): A robust, open-source python library for materials analysis}.
\newblock \emph{Computational Materials Science}, 68: 314--319.

\bibitem[{Pan et~al.(2021)Pan, Ganose, Horton, Aykol, Persson, Zimmermann, and Jain}]{Pan2021BenchmarkingStructures}
Pan, H.; Ganose, A.~M.; Horton, M.; Aykol, M.; Persson, K.~A.; Zimmermann, N.~E.; and Jain, A. 2021.
\newblock {Benchmarking Coordination Number Prediction Algorithms on Inorganic Crystal Structures}.
\newblock \emph{Inorganic Chemistry}, 60(3): 1590--1603.

\bibitem[{Riaz et~al.(2014)Riaz, Jung, Chang, Shin, and Lee}]{Riaz2014Carbon-Arrays}
Riaz, A.; Jung, K.-N.; Chang, W.; Shin, K.-H.; and Lee, J.-W. 2014.
\newblock {Carbon-, Binder-, and Precious Metal-Free Cathodes for Non-Aqueous Lithium–Oxygen Batteries: Nanoflake-Decorated Nanoneedle Oxide Arrays}.
\newblock \emph{ACS Applied Materials {\&} Interfaces}, 6(20): 17815--17822.

\bibitem[{Sendek et~al.(2017)Sendek, Yang, Cubuk, Duerloo, Cui, and Reed}]{Sendek2017HolisticMaterials}
Sendek, A.~D.; Yang, Q.; Cubuk, E.~D.; Duerloo, K.-A.~N.; Cui, Y.; and Reed, E.~J. 2017.
\newblock {Holistic computational structure screening of more than 12000 candidates for solid lithium-ion conductor materials}.
\newblock \emph{Energy {\&} Environmental Science}, 10(1): 306--320.

\bibitem[{Shannon(1948)}]{Shannon1948ACommunication}
Shannon, C.~E. 1948.
\newblock {A Mathematical Theory of Communication}.
\newblock \emph{Bell System Technical Journal}, 27(3): 379--423.

\bibitem[{{Sherif Abdulkader Tawfik}(2023)}]{SherifAbdulkaderTawfik2023OganessonHttps://github.com/sheriftawfikabbas/oganesson}
{Sherif Abdulkader Tawfik}. 2023.
\newblock {Oganesson https://github.com/sheriftawfikabbas/oganesson}.

\bibitem[{Wang, Botti, and Marques(2021)}]{Wang2021PredictingSimilarity}
Wang, H.~C.; Botti, S.; and Marques, M.~A. 2021.
\newblock {Predicting stable crystalline compounds using chemical similarity}.
\newblock \emph{npj Computational Materials 2021 7:1}, 7(1): 1--9.

\bibitem[{Ward et~al.(2016)Ward, Agrawal, Choudhary, and Wolverton}]{Ward2016AMaterials}
Ward, L.; Agrawal, A.; Choudhary, A.; and Wolverton, C. 2016.
\newblock {A general-purpose machine learning framework for predicting properties of inorganic materials}.
\newblock \emph{npj Computational Materials}, 2(1): 16028.

\bibitem[{Ward et~al.(2018)Ward, Dunn, Faghaninia, Zimmermann, Bajaj, Wang, Montoya, Chen, Bystrom, Dylla, Chard, Asta, Persson, Snyder, Foster, and Jain}]{Ward2018Matminer:Mining}
Ward, L.; Dunn, A.; Faghaninia, A.; Zimmermann, N.~E.; Bajaj, S.; Wang, Q.; Montoya, J.; Chen, J.; Bystrom, K.; Dylla, M.; Chard, K.; Asta, M.; Persson, K.~A.; Snyder, G.~J.; Foster, I.; and Jain, A. 2018.
\newblock {Matminer: An open source toolkit for materials data mining}.
\newblock \emph{Computational Materials Science}, 152: 60--69.

\bibitem[{Wei et~al.(2022)Wei, Fu, Siriwardane, Yang, Omee, Dong, Xin, and Hu}]{Wei2022TCSP:Discovery}
Wei, L.; Fu, N.; Siriwardane, E.~M.; Yang, W.; Omee, S.~S.; Dong, R.; Xin, R.; and Hu, J. 2022.
\newblock {TCSP: A Template-Based Crystal Structure Prediction Algorithm for Materials Discovery}.
\newblock \emph{Inorganic Chemistry}, 61(22): 8431--8439.

\bibitem[{Wen~DeepMind et~al.(2020)Wen~DeepMind, Precup~DeepMind, Ibrahimi~DeepMind, Barreto~DeepMind, Van Roy~DeepMind, and Singh~DeepMind}]{WenDeepMind2020OnLearning}
Wen~DeepMind, Z.; Precup~DeepMind, D.; Ibrahimi~DeepMind, M.; Barreto~DeepMind, A.; Van Roy~DeepMind, B.; and Singh~DeepMind, S. 2020.
\newblock {On Efficiency in Hierarchical Reinforcement Learning}.
\newblock In \emph{Conference on Neural Information Processing Systems}.

\bibitem[{Xie et~al.(2022)Xie, Fu, Ganea, Barzilay, and Jaakkola}]{Xie2022CrystalGeneration}
Xie, T.; Fu, X.; Ganea, O.-E.; Barzilay, R.; and Jaakkola, T. 2022.
\newblock {Crystal Diffusion Variational Autoencoder for Periodic Material Generation}.
\newblock In \emph{Proceeding of International Conference on Learning Representations}.

\bibitem[{Yan et~al.(2022)Yan, Liu, Lin, and Ji}]{Yan2022PeriodicPrediction}
Yan, K.; Liu, Y.; Lin, Y.; and Ji, S. 2022.
\newblock {Periodic Graph Transformers for Crystal Material Property Prediction}.
\newblock In \emph{Proceedings of Advances in Neural Information Processing Systems}.

\bibitem[{Zamaraeva et~al.(2023)Zamaraeva, Collins, Antypov, Gusev, Savani, Dyer, Darling, Potapov, Rosseinsky, and Spirakis}]{Zamaraeva2023ReinforcementPrediction}
Zamaraeva, E.; Collins, C.~M.; Antypov, D.; Gusev, V.~V.; Savani, R.; Dyer, M.~S.; Darling, G.~R.; Potapov, I.; Rosseinsky, M.~J.; and Spirakis, P.~G. 2023.
\newblock {Reinforcement Learning in Crystal Structure Prediction}.
\newblock \emph{Digital Discovery}.

\bibitem[{Zhang et~al.(2022)Zhang, Li, Yuan, Wu, Dou, and Han}]{Zhang2022MgAlBatteries}
Zhang, J.; Li, T.; Yuan, Q.; Wu, Y.; Dou, Y.; and Han, J. 2022.
\newblock {MgAl Saponite as a Transition-Metal-Free Anode Material for Lithium-Ion Batteries}.
\newblock \emph{ACS Applied Materials {\&} Interfaces}, 14(49): 54812--54821.

\bibitem[{Zhao et~al.(2021)Zhao, Al-Fahdi, Hu, D~Siriwardane, Song, Nasiri, Hu, Zhao, D~Siriwardane, Song, Nasiri, Hu, Al-Fahdi, and Hu}]{Zhao2021High-ThroughputNetworks}
Zhao, Y.; Al-Fahdi, M.; Hu, M.; D~Siriwardane, E.~M.; Song, Y.; Nasiri, A.; Hu, J.; Zhao, Y.; D~Siriwardane, E.~M.; Song, Y.; Nasiri, A.; Hu, J.; Al-Fahdi, M.; and Hu, M. 2021.
\newblock {High-Throughput Discovery of Novel Cubic Crystal Materials Using Deep Generative Neural Networks}.
\newblock \emph{Advanced Science}, 8(20): 2100566.

\bibitem[{Zhao et~al.(2023)Zhao, Siriwardane, Wu, Fu, Al-Fahdi, Hu, and Hu}]{Zhao2023PhysicsConstraints}
Zhao, Y.; Siriwardane, E.~M.; Wu, Z.; Fu, N.; Al-Fahdi, M.; Hu, M.; and Hu, J. 2023.
\newblock {Physics guided deep learning for generative design of crystal materials with symmetry constraints}.
\newblock \emph{npj Computational Materials 2023 9:1}, 9(1): 1--12.

\end{thebibliography}
\onecolumn

\section{Geometrical Characteristics}
\label{sec:lattice_spacegroup_character}
The list of geometrical characteristics of space groups is provided in the Table \ref{stab:lattice_spacegroup}.
\begin{table*}[ht]
\centering
\caption{Geometrical characteristics of space groups in terms of lattice angles or lengths}
\label{stab:lattice_spacegroup}
\begin{tabular}{|l|l|l|l|l|}
\hline
Space group & Crystal family & Lengths constraints & Angles constraints & Parameters search space\\ \hline
1-2         & Triclinic c    & None                & None               & $a, b, c, \alpha, \beta, \gamma$ \\ \hline
3-15        & Monoclinic     & None                & $\alpha = \beta = 90^\circ$ & $a, b, c, \gamma$\\ \hline
16-74       & Orthorhombic   & None                & $\alpha = \beta = \gamma =  90^\circ$ & a, b, c\\ \hline
75-142      & Tetragonal     & $a = b$             & $\alpha = \beta = \gamma =  90^\circ$ & a, c\\ \hline
143-194     & Hexagonal      & $a = b$             & $\alpha = \beta =  90^\circ, \gamma =  120^\circ$ & a, c\\ \hline
195-230     & Cubic          & $a = b = c$         & $\alpha = \beta = \gamma =  90^\circ$ & a\\ \hline
\end{tabular}
\end{table*}
\section{Continous GFlownet assumptions}
The soundness of the theory of continuous GFlownet relies on a set of assumptions:
\begin{itemize}
    \item The structure of the state space must allow all states to be reachable from the source state $s_0$.
    \item The structure must ensure that the number of steps required to reach any state from $s_0$ is bounded.
    \item The learned probability measures need to be expressed through densities over states, rather than over actions.
\end{itemize}

Our crystal structure state space and training framework satisfy these assumptions as:
\begin{itemize}
    \item Given the $s_0$ as the empty crystal structure with space group 1 (lowest symmetry without any constraints on the lattice parameters and no symmetry operation) and initial lattice params, the measurable pointed graph is defined as $\mathcal{S}={s_0}\cup [min_l,max_l]^3\cup [min_a, max_a]^3 \cup [0,1]^T$ where $min_l$, $max_l$ are minimum and maximum lattice length, $min_a,max_a$  are minimum and maximum lattice angle.
    \item We have the upper limit $T$ for the trajectory length during the trajectory sampling process. Therefore the number of steps to reach any state is bounded by $T$.
    \item The forward $p_F$ and backward $p_B$ policy networks learn the distribution over states $P_F(s^t_{al}|s^{t-1}_{sg},s^{t-1}_{al},s^t_{sg})$, $P_F(s^t_{sg}|s^{t-1}_{sg},s^{t-1}_{al})$, $P_B(s^{t-1}_{al}|s^{t}_{sg},s^{t}_{al},s^{t-1}_{sg})$, $P_B(s^{t-1}_{sg}|s^{t}_{sg},s^{t}_{al})$.
\end{itemize}
\section{Battery material discovery task}
\label{sec:batterydiscovery}
Discovering stable material is one of the main targets of machine learning in material science. Finding new battery materials with improved properties such as higher voltage cathodes and higher diffusivity solid-state electrolytes is one of the key challenges in material science \cite{Muy2019High-ThroughputDescriptors,Kahle2020High-throughputConductors}. We were motivated to introduce a generative approach for battery material discovery which we expect to unravel novel material compositions and structures that are not accessible to high-throughput screening approaches.

Discovering battery material is crucial in many aspects such as safety, environment, and cost. Different from stable material discovery, battery materials are also required to satisfy constraints dedicated to different types of batteries. For example, light-weight, transition-metal-free cation battery materials require a subset of elements. Another example is designing high ionic mobility battery material requires porous structures with low band gap. Such complicated constraints require a flexible model. Our model offers an easy tool to incorporate domain knowledge and requirements into the crystal structure sampling process.

Battery materials, such as cathodes and solid-state electrolytes, can potentially include metallic and heavy elements, even rare-earth elements such as the LLTZ solid-state electrolyte. However, our motivation in this work is to design battery materials composed of Earth-abundant, non-metal elements. There is a large number of research publications that aim at substituting traditional battery materials with metal-free alternatives:
\begin{itemize}
    \item An anode material that is transition metal-free \cite{Zhang2022MgAlBatteries}
    \item A cathode material that is precious metal-free \cite{Riaz2014Carbon-Arrays}
    \item A transition metal-free cathode \cite{Kim2014NovelBatteries}
\end{itemize}
In our work, we have limited our attention to the named elements having the highest Earth-abundance according to Table S1 in the seminal work \cite{Sendek2017HolisticMaterials}. We have excluded abundant elements such as Mg and Ca, to optimize the computational cost.

\section{Implementation Details}
\label{sec:implementdetail}
\subsection{State Graph Construction}
We determine the edges of the crystal structure graph using k-nearest neighbor atoms within  $4$ \r{A}. The node feature is the coordinate and the atomic number of the atom. 
\subsection{MatErials Graph Network (MEGNet)}
\label{sec:megnet}
MEGNet \cite{Chen2019GraphCrystals} is a graph neural network that uses two-body distance as edge features. Given the input graph $\mathcal{G}=(\mathcal{V},\mathcal{E},u)$ where $\mathcal{X}$ are atoms, $\mathcal{E}$ is edges, and $u$ is global state attribute which is, for example, temperature. In our implementation, the global state attribute is set as $0.0$.

First, the bond's attribute $(e_k, r_k, s_k)$ where $e_k$ is the bond feature, $r_k$ and $s_k$ are atoms indices of the bond, are updated as:
\begin{equation}
    e_k'=\phi_e [v_{s_k};v_{r_k};e_k;u]
\end{equation}
where $\phi_e$ is the update function and $[;]$ is concatenation.

Second, the atom feature is updated as:
\begin{equation}
    \hat{v_ie} = \frac{1}{N^e_i} \sum_{k=1}^{N^e_i}\{e'_k\}_{r_k=1}
\end{equation}
\begin{equation}
    v_i' = \phi_v [\hat{v_ie});v+i;u)
\end{equation}
The update functions $\phi_v, \phi_e$ are the MLP layers.

\subsection{Graph Convolution Neural Network (GCN)}
\label{sec:gcnmp}
GCN \cite{Kipf2017Semi-supervisedNetworks} is a convolutional network designed to learn the node-level representation of graph structure $\mathcal{G}=(\mathcal{X},\mathcal{E})$, where $\mathcal{X}$
is the node feature matrix of $N$ nodes and $\mathcal{E}\in R^{N\times N}$
is the adjacency matrix. Given $W^{l}$
be the weight matrix at $l$-th layer, the graph convolution operation
is then defined as:

\begin{align}
H^{1} & =\mathcal{X},\\
H^{l} & =\sigma\left(\tilde{D}^{-\frac{1}{2}}\tilde{\mathcal{E}}\tilde{D}^{-\frac{1}{2}}H^{l-1}W^{l-1}\right),\label{eq:gcn_layer}
\end{align}
where $\mathcal{\tilde{E}=\mathcal{E}+I}$ is the adjacency matrix
with self-loop in each node. $\mathcal{I}$ is the identity matrix
and $\tilde{D}=\sum_{j}\mathcal{\tilde{E}}_{ij}$ and $\sigma$ is
a non-linear function.

\subsection{Choosing properties to optimize in the reward function}
\label{sec:motivationrewardfunction}
Formation energy, bond length, density, and validity are chosen because they are direct and fundamental properties to be used as the reward function for stable crystal structure generation. Training on these properties allows the model to be universal and flexible enough to be further finetuned to adapt to other tasks or target properties such as band gap.

Some physical interactions such as electronegativity and atomic radius can influence the atoms' position. However, their influence on stability can be indirect and complex. For example, two atoms can still form a bond regardless of the absolute value of the difference in electronegativity ($\Delta EN$) (covalent bond or ionic bond). Atom radius has four common definitions: Van der Waals radius, ionic radius, metallic radius, and covalent radius. Choosing the radius type also depends on the bond types. On the other hand, bond length is a more straightforward and universal measurement.

Terms such as energy above hull $E_{hull}$ which is very informative and directly linked to the stability of the structure. Unfortunately, the computation cost $E_{hull}$ is too high (requires relaxing the structure and then getting the energy of the final state to compute the phase diagram) even using a deep learning model like M3GNet. Therefore, our choice of reward set function also considers balancing between the computation efficiency and fundamentals.

\subsection{Crystal Structure Density}
\label{sec:density}
The generated crystal structure density is defined as the ratio of mass $m$ of the unit cell  over the volume $V$ of unit cell:
\begin{equation}
    P(x) = \frac{m}{V}
\end{equation}
In our implementation, the density is calcuated using pymatgen library.\citep{Ong2013PythonAnalysis}.

\subsection{Diversity Metrics}
\label{sec:diversity}
Structure diversity is computed based on the CrystalNNFingerprint (CNN fingerprint) \cite{Ward2018Matminer:Mining}. CrystalNNFingerprint computes the fingerprint of a given site $i$ using its coordination features and neighbors. The site $i$ neighbors are determined by the CrystalNN neighbor-finding algorithm. The fingerprint of a crystal structure is the average of fingerprints of all sites. 

Composition fingerprint is computed using the statistics of Magpie, computed by element stoichiometry \cite{Ward2018Matminer:Mining,Ward2016AMaterials}. We use ElementProperty.from\_preset('magpie') in Matminer as material composition fingerprint.

Then we define diversity as:
\begin{equation}
    Diversity = \frac{1}{n}\sum_{i,j\in N_{gen}}d(f_{fp}(i),f_{fp}(j))
\end{equation}
where $N_{gen}$ is the set generated crystal structures, $n$ is the number of structure in set $N_{gen}$. In our experiment, $N_{gen}$ is the top-K crystal structures ranked by reward, $f_{fp}$ is the structure fingerprint in case of structure diversity or composition fingerprint in case of composition diversity, $d$ is the Euclidean distance.

\subsection{Unique Mode}
\label{sec:mode}
We define a unique \textit{mode} if it satisfies four conditions. The first condition is three types of validity defined in Sec. 5.1. The second condition is that the crystal structure satisfies minimum distance between any two atoms constraints. The third condition is that the structure must have negative formation energy. The fourth condition is that the composition cannot be the same as other modes.

\subsection{Formation Energy Prediction}
\label{sec:formemodel}
We use M3GNet \citep{Chen2022ATable} to predict the formation energy. As the M3GNet is only trained on the Material Project valid crystal structure, the predictions for invalid structures may be inaccurate and have abnormally low or high formation energy. As our formation energy score function is an exponential function (Sec. \ref{sec:reward}), the formation energy term can get really large, affecting other terms. Therefore, we clamp the formation energy prediction value from -10.0 to 10.0 eV/atom.

\subsection{Hyper-parameters and experiment set up}
\label{sec:hyperparam}
The hyper-parameter used for our proposed method is presented in Tab. \ref{stab:hyparam}. We decide the scaler weights based on the ablation study (Table \ref{tab:matchM3GNet_ablation}), values over epochs (Fig. \ref{fig:term_ep}), and our intuition. In the ablation study, without the bond score term, the result drops significantly. Because the bond score influences the generation quality heavily, we set the weight $w_b$ as $0.5$. Both the density and formation energy affect the results at the same level. Therefore, we set both weights $w_e$ and $w_p$ as $0.2$. The value of validity over epochs (Figure \ref{fig:term_ep}d) converges quickly close to 1.0 which is the maximum value. It seems that validity is an easy term for the model to learn compared to density and formation energy. Therefore, we lower the validity weight $w_c$ to $0.1$.

We run our experiments in Tab. \ref{Tab1:valid_3}-\ref{tab:matchM3GNet} in the main text three times with random seeds. The result for CDVAE in Table \ref{tab:general_chem_space} in the main text is reported in their paper.

\begin{table}[h]
\caption{Hyper-parameters}
\label{stab:hyparam}
\begin{center}
\begin{tabular}{|l|l|}
\hline
Hyper-paramters & Value \\ \hline
Learning rate         & 0.001 \\ \hline
Learning rate Z        & 0.1\\ \hline
Optimizer & Adam \\ \hline
Learning rate scheduler $\gamma$ & 1.0 \\ \hline
Initial logZ       & 0.0\\ \hline
Batch size      &  32\\ \hline
$w_e$ (Eq. \ref{eq:reward-func})     & 0.2\\ \hline
$w_p$ (Eq. \ref{eq:reward-func})     & 0.2\\ \hline
$w_b$ (Eq. \ref{eq:reward-func})    & 0.5\\ \hline
$w_c$ (Eq. \ref{eq:reward-func})    & 0.1\\ \hline
\end{tabular}
\end{center}
\end{table}

\subsection{MP-Battery dataset}
\label{sec:mpbatterydataset}
To address the requirements of the battery material discovery task outlined in Sec. \ref{sec:battery} and Appendix Sec. \ref{sec:batterydiscovery}, we developed the MP-Battery dataset using the Materials Project database \citep{Jain2013Commentary:Innovation}. We specifically selected materials from the 187,687 available in the dataset that contain light elements such as Be, B, C, N, O, Si, P, S, and Cl, alongside one of the alkali metals Li, Na, or K. This process resulted in the creation of the MP-Battery dataset, which includes 2,060 distinct materials. The MP-Battery dataset is included in the Supplemental data with train/validation/test split.

\section{DFT Validation}
We sample and choose the top-K generated materials for DFT verification. Out of 100 crystal structures, 95\% (95) are successfully optimized. The complete list of DFT-optimized structures is included in the Supplement data. The distribution of total energy per atom
of DFT-optimized structures shows that all DFT-
optimized structures have total energy lower than 0 eV/atom. 

\begin{figure}
\begin{centering}
{\includegraphics[width=0.8\textwidth]{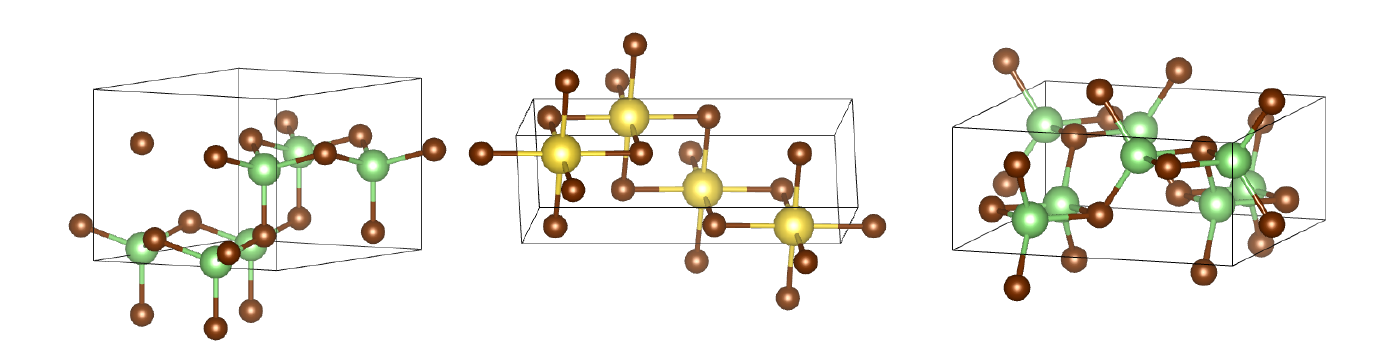}}
\par\end{centering}
\caption{The examples of DFT-verified stable structures ($E_{hull}=0$ eV/atom).} \label{fig:Ehull_stable}
\end{figure}

For the structures that were optimized using DFT, we have computed the crystal decomposition energy ($E_D$) for each of the structures using the code available in the Python package oganesson \citep{SherifAbdulkaderTawfik2023OganessonHttps://github.com/sheriftawfikabbas/oganesson}. $E_D$ is calculated by obtaining the difference in energy between the structure's energy and the energies of competing phases that are available in Materials Project \citep{Jain2013Commentary:Innovation}. For metastable structures (where $E_D$ values are negative), the energy above the convex hull ($E_{hull}$) is equivalent to $-E_{D}$. Structures that do not decompose into competing phases will have an $E_H = 0$, and they are known as ground state phases. Out of 100 crystal structures, 7\% structure are stable ($E_{hull}=0 eV/atom$) and 56\% structure are metastable ($E_{hull} <0.05 eV/atom$). The examples of DFT-verified stable structures ($E_{hull}=0$ eV/atom) are presented in Fig. \ref{fig:Ehull_stable}. 

\begin{figure}
\begin{centering}
{\includegraphics[width=0.6\textwidth]{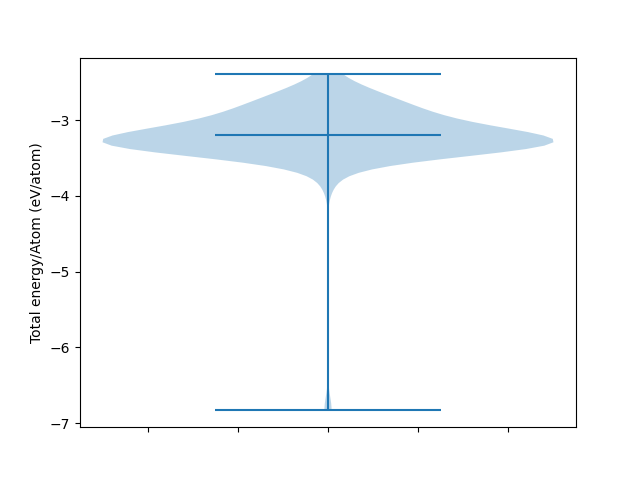}}
\par\end{centering}
\caption{The distribution of total energy per atom
of DFT-optimized structures.} \label{fig:DFTenergy_dist}
\end{figure}

\section{Reward and Rewards Terms over Epochs}
\begin{figure}[h]
\centering     
\subfigure[Average bond score over epochs]{\label{fig:bs_ep}\includegraphics[width=0.4\textwidth]{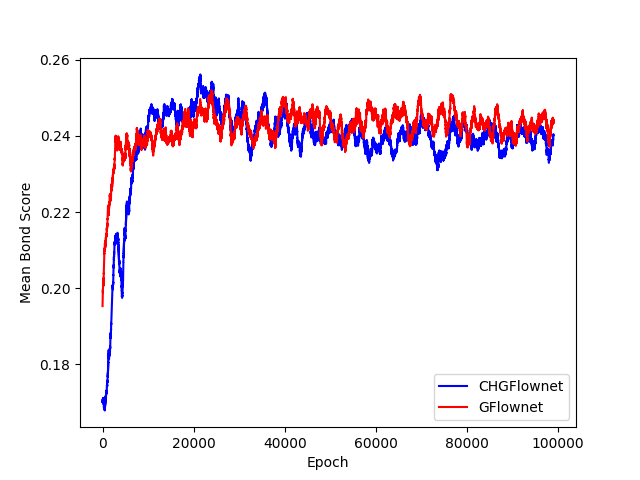}}
\subfigure[Average formation energy score over epochs]{\label{fig:es_ep}\includegraphics[width=0.4\textwidth]{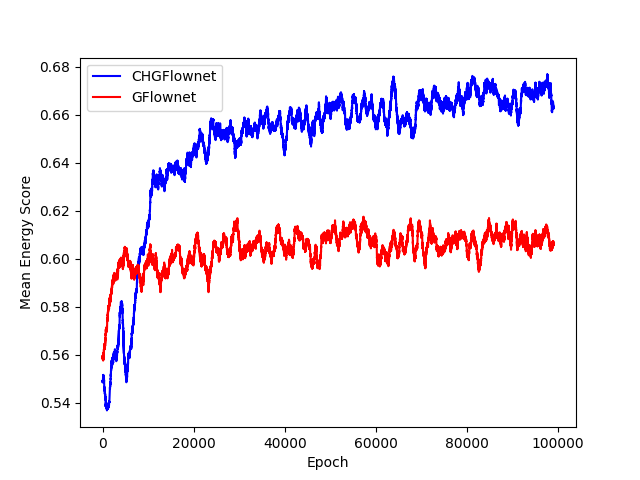}}
\subfigure[Average density score over epochs]{\label{fig:ds_ep}\includegraphics[width=0.4\textwidth]{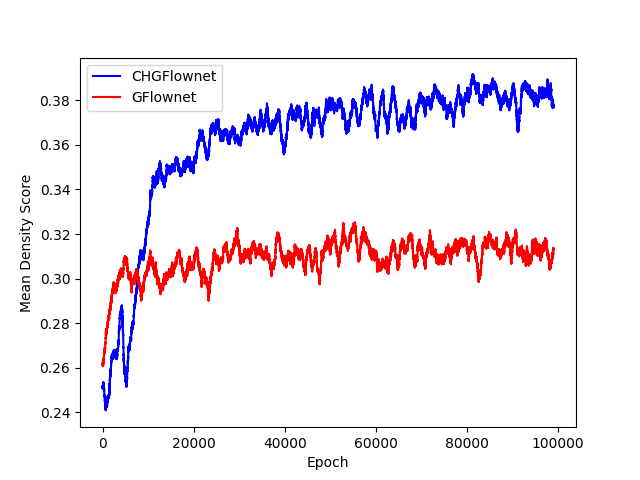}}
\subfigure[Average validity score over epochs]{\label{fig:vs_ep}\includegraphics[width=0.4\textwidth]{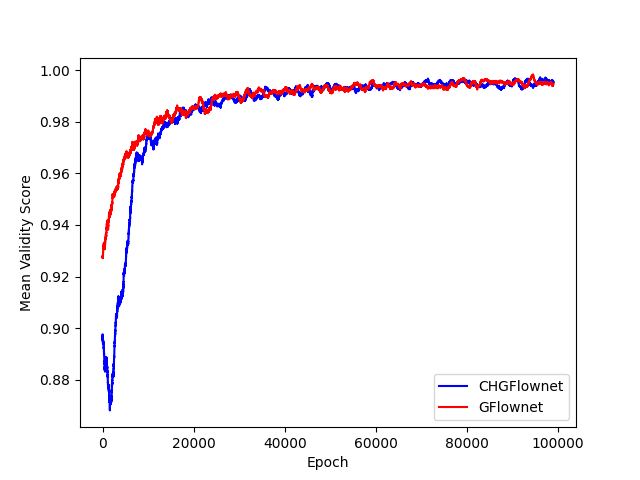}}
\caption{Average of terms of reward function over $10^5$ epochs.}
\label{fig:term_ep}
\end{figure}

During the training process, we monitor four terms of the reward function: bond score, formation energy, density, and validity score term. The average score terms of each sampled batch over epochs are plotted in Fig. \ref{fig:term_ep}.

\section{Distribution of reward terms}
\begin{figure}[h!]
\centering     
\subfigure[Distribution of formation energy term]{\label{fig:bs_ep_dist}\includegraphics[width=0.4\textwidth]{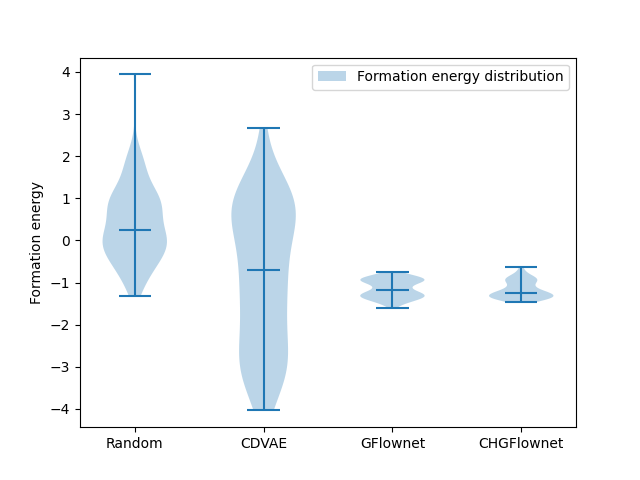}}
\subfigure[Distribution of density term]{\label{fig:es_ep_dist}\includegraphics[width=0.4\textwidth]{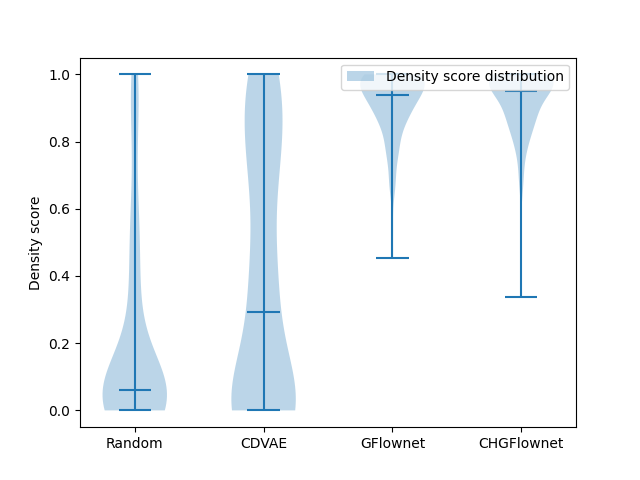}}
\subfigure[Distribution of bond score term]{\label{fig:ds_ep_dist}\includegraphics[width=0.4\textwidth]{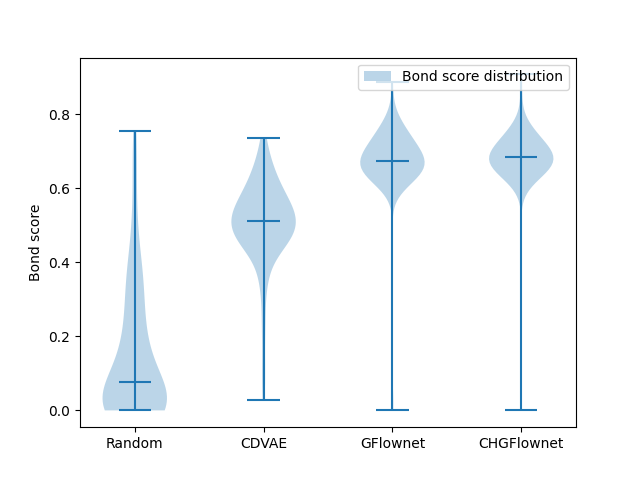}}
\caption{The distribution of structures sampled untrained SHAFT, by CDVAE, GFlownet, and SHAFT regarding terms in the reward function.}
\label{fig:term_ep_dist}
\end{figure}

The distribution of structures sampled by untrained SHAFT, CDVAE, trained GFlownet, and trained SHAFT regarding terms in the reward function are plotted in Fig.\ref{fig:term_ep_dist}. We can see a distribution shift between the trained and untrained SHAFT. Without any training, the sampled structure can be considered random.

\section{Stability evaluated by energy above hull ($E_{hull}$)}
The stability of the generated crystal structure can also be evaluated by using the phase diagram to perform convex hull analysis. Convex hull analysis shows the stability of the generated structure composition within the composition space. The energy above hull ($E_{hull}$) indicates the decomposition energy of the generated composition into the combination of the stable phases. Higher $E_{hull}$ indicates less thermodynamic stability.

Because relaxing structure and calculating energy using DFT is computationally expensive, we can approximate the $E_{hull}$ value using a deep learning model. Specifically, we relax the generated crystal structure using M3GNet Relax tool \citep{Chen2022ATable} and then estimate the energy of the final structure using M3GNet \citep{Chen2022ATable}. The generated structure's composition and estimated energy are used for convex hull analysis using Pymatgen \citep{Ong2013PythonAnalysis}. The average of energy above the hull is shown in Tab. \ref{tab:ehull}.

\begin{table}[h]
\caption{Average energy above hull $E_{hull}$ estimated using M3GNet \citep{Chen2022ATable} and Pymatgen \citep{Ong2013PythonAnalysis}}
\label{tab:ehull}
\begin{center}
\begin{tabular}{ll}
\toprule
Methods & $E_{hull}$ $\downarrow$ \\
\hline
CDVAE & 0.81 $\pm0.03$\\
GFlowNet & 0.32 $\pm0.02$\\
SHAFT & 0.29 $\pm0.03$\\
\bottomrule
\end{tabular}
\end{center}
\end{table}

\section{Generating stable structure with target property}
Our proposed framework SHAFT offers the flexibility to optimize multiple objectives via the reward function. We experiment to generate stable structures with low band gap. The experiment setup is similar to the battery material discovery task (Sec.\ref{sec:exp}). Material with low band gap is advantageous for battery material as it has high ion mobility. We add the following band gap term for the near zero band gap constraint:
\begin{equation}
    R_{bg}(x)= a_{bg}e^{\frac{-(P_{bg}(x)-b_{bg})^2}{2c_{bg}^2}}
\end{equation}
where $a_{bg}=3.0$, $b_{bg} = 0.0$, $c_{bg} = 0.5$.
Then the reward function is defined as:
\begin{equation}
    \label{eq:reward-func_bg}
    R(x) = w_eR_e(x) + w_{bg}R_{bg}(x)+ w_pR_P(x) + w_bR_{bond}(x) + w_cR_{comp}(x)
\end{equation}
We set weights as $w_e=w_{bg}=0.1$, $w_p=0.1$, $w_b=0.5$, and $w_c=0.1$. To demonstrate that the generated structures are stable and have near zero band gap, we present the average of formation and band gap in Tab.\ref{Tab:forme_bandgap}.
\begin{table}[h]
\caption{Average of formation energy and band gap of the generated structures. We evaluate the top-K crystal structures ranked by the reward function after $8\times10^5$ states visited.}
\label{Tab:forme_bandgap}
\begin{center}
\begin{tabular}{lccr}
\hline
Method & Energy $\downarrow$ & \% energy $< 0$ $\uparrow$ & Band gap $\downarrow$\\ \hline
CDVAE & -0.82    & 83.33 & 0.45\\
GFlowNet &-0.63&90.91& 0.08\\
SHAFT &-0.84&100.00 & 0.04\\
\hline
\end{tabular}
\end{center}
\end{table}

\section{Limitations}
\label{sec:limitation}
One of the advantages of GFlownet is applying hard constraints directly during the sampling process. It will be more efficient to apply the bond length constraints while sampling the atom's coordinate instead of using soft constraint in the reward function. Additionally, further atoms' coordinate refinement using diffusion allows more stable structures.  

\end{document}